\documentclass[11pt,a4paper]{article}
\usepackage[hiresbb]{graphicx}
\usepackage{amssymb}
\usepackage{amsmath}
\usepackage{apacite}
\usepackage{authblk}
\usepackage{url}
\usepackage{anysize}
\usepackage{pdfsync}

\marginsize{2.5cm}{2.5cm}{2.5cm}{2.5cm}

\newcommand{\citep}[1]{\shortcite{#1}} 
\newcommand{\citet}[1]{\shortciteA{#1}} 

\newcommand{\conf}[2]{#1--#2}

\title{Higher coordination with less control -- A result of information
maximization in the sensorimotor loop}

\author[1*]{Keyan Zahedi}
\author[1,2]{Nihat Ay}
\author[1]{Ralf Der}
\affil[1]{MPI for Mathematics in the Sciences, Inselstrasse 22, 04103 Leipzig, Germany}
\affil[2]{Santa Fe Institute, 1399 Hyde Park Road, Santa Fe, NM 8501, USA}
\affil[*]{Corresponding author, Email: zahedi@mis.mpg.de, Phone: +49 (0) 341
9959 544, Fax: +49 (0) 341 9959 555}

\begin{document}
\maketitle

\newpage

\begin{abstract}
  This work presents a novel learning method in the context of embodied
  artificial intelligence and self-organization, which has as few assumptions
  and restrictions as possible about the world and the underlying model. The learning rule is
  derived from the principle of maximizing the predictive information in the
  sensorimotor loop. It is evaluated on robot chains of varying length with
  individually controlled, non-communicating segments.
  The comparison of the results shows that maximizing the
  predictive information per wheel leads to a higher coordinated behavior of
  the physically connected robots compared to a maximization per robot.
  Another focus of this paper is the analysis of the effect of the robot chain
  length on the overall behavior of the robots. It will be shown that longer
  chains with less capable controllers outperform those of shorter length and
  more complex controllers. The reason is found and discussed in the
  information-geometric interpretation of the learning process.
\end{abstract}

\newpage

\section{Introduction}
\label{sec:introduction}
An ongoing research topic is to understand the learning and adaptation processes of
cognitive systems. This paper will not define what a cognitive system is.
Instead, the term \emph{cognitive system} is used as an abstract concept just as
in \citep{Brooks1991Intelligence-Without-Reason}. Generally speaking, in this
context, cognition is understood as a process that transforms sensory data into
motor commands using some form of internal (non-symbolic) representation
\citep{Forster1993Wissen-und-Gewissen}. Hence, cognition is a process which
lives in the sensorimotor loop \citep{Cliff1990Computational-neuroethology:-a},
or, otherwise stated, to understand adaptation and learning it is essential to
take the body and environment into account \citep{Pfeifer2006How-the-Body}. In
order to investigate learning and adaptation, this work follows the approach of
embodied artificial intelligence, first described by
\citet{Brooks1986A-Robust-Layered} and later refined by
\citet{Pfeifer2006How-the-Body}, in which complete robotic systems of lower
complexity are built and understood before the complexity is then gradually
increased.

In the field of embodied artificial intelligence, learning and adaptation rules
are very often specific in either their possible applications or applicable
models (with few exceptions, such as
\citep{Pasemann2004Adaptive-Behaviour-Control}). Examples are the ISO/ICO
learning \citep{Porr2003Sequence-Learning-in-a} which is limited to a single
neuron network, and the Homeostatic learning rule by
\citet{Di-Paolo2000Homeostatic-adaptation-to}, which operates on a fully
connected recurrent neural network, but requires a trigger mechanism. We are
interested in a first principle learning rule, i.e.~a learning rule which is
independent of the model structure and requires as few assumptions as possible
about the morphology
and environment. This sounds like a contradiction to the statement of the first
paragraph (the sensorimotor loop is essential to understand cognition), and
therefore, needs to be elaborated. We are looking at a very basic level of
unsupervised and self-organized learning, i.e.~before any task-dependent
learning occurs. The question is, how can a system, with no knowledge of itself
or the world, learn enough to perform coordinated interactions within the
environment? An analogy is an infant who performs what is known as \emph{motor-}
or \emph{body babbling} in order to learn how to produce facial expressions
\citep{Meltzoff1997Explaining-facial-imitation:}.
\citet{Clark1996Being-There:-Putting} describes it more generally, and states
that there is evidence by \citet{Thelen1996A-dynamic-systems}, that infants
learn about the world through actions (and that the acquired knowledge is also
action-specific). 
It is this form of learning that we are interested in.

We can
now reformulate and specify the question above as the question of how a system
may interactively gain maximal information about itself and the world.
This is related to other self-motivated learning
approaches. An example, and probably the first implementation is \citep{Schmidhuber1990A-possibility-for}, in which,
additionally to a controller network, a model network is adapted and the
prediction error of the latter is used as a reinforcement signal to the former. A similar
approach, but with a very different architecture, is used by
\citet{Oudeyer2007Intrinsic-Motivation-Systems} and \citet{Kaplan2004Maximizing-Learning-Progress:},
who use the learning progress (a function of the prediction
error), as a reinforcement signal.
\citet{Barto2004Intrinsically-motivated-learning} uses the prediction error of
skill models to build hierarchical skill collections.
Two further approaches are discussed by 
\citet{Schmidhuber2009Driven-by-Compression} and
\citeauthor{Steels2004The-Autotelic-Principle} \citeyear{Steels2004The-Autotelic-Principle,Steels2007Scaffolding-Language-Emergence}. The former proposes the utilization
of the compression progression of a system as its reinforcement signal, while the
latter proposes the \emph{Autotelic Principle}, i.e.~the balance of skill and
challenge of behavioral components as the motivation for open ended development.
The approach proposed here is probably most similar to the work of
\citet{Storck1995Reinforcement-Driven-Information}, in which
the difference of consecutive probability distributions of the world model, measured
e.g.~by the Kullback–-Leibler divergence, is used as a positive reward for the
reinforcement learning.

All mentioned approaches use intrinsically generated reinforcement signals as an
input to a learning algorithm. The main difference here is that we do not use
reinforcement learning in the sense that the predictive information is used as a
reward function. Instead, we directly calculate the gradient of the policy as a
result of the current locally available approximation of the predictive
information. Nevertheless, a
function of the predictive information, as proposed in this work, can be used as
a reinforcement signal in all of the approaches mentioned above.

Posing the question in the form given above, it is natural to propose Shannon's information
theory \citep{Shannon1948A-mathematical-theory} as the foundation to formulate
such a first principle learning rule. There are good reasons to assume
information maximization as a guiding principle in cognitive systems.
\citet{Linsker1988Self-organization-in-a} reproduced receptive fields, similar
to those of the visual cortex, by applying the InfoMax principle to a
feed-forward neural network, i.e.~a neural network in which earlier layers
maximize the information passed to the next layers. A similar principle was
also shown experimentally for single neuron recordings by
\citet{Laughlin1981A-simple-coding}. Unfortunately, models in this context are
again limited by underlying network or model structures (e.g.~a feed-forward
network with localized and highly symmetric connectivity in Linsker's case). In
addition, information theory has been applied to the sensorimotor loop by
e.g.~\citet{Lungarella2005Information-Self-Structuring:-Key} and
\citet{Polani2006Relevant-Information-in}. The former publication investigated
different information theoretic measures in the sensorimotor loop, while the
latter asked the question of what is the minimum amount of information that is
required by an agent in order to maximize a utility function.

Following this introductory section, the next section first introduces the
sensorimotor loop in the context of information theory, followed by the
derivation of the learning rule. The third section presents different
experiments based on chains of robots on which the learning rule was
implemented. The fourth section discusses the results, and the last section
concludes.

\section{Learning Rule}
\label{sec:learning rule}
This work presents a learning rule in the context of embodied artificial
intelligence, based on the InfoMax principle. Hence, in order to formulate such
a learning rule, it is necessary to define the sensorimotor loop in the context
of information theory. This is done in the following paragraphs.

\begin{figure}[h]
  \begin{center}
    \includegraphics[width=16cm]{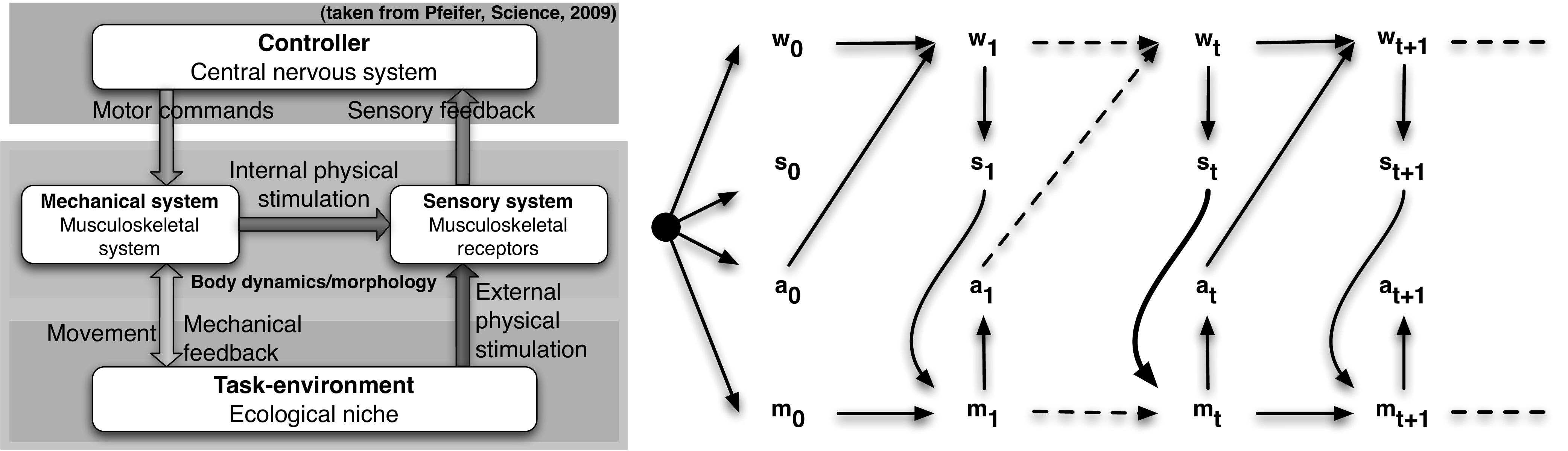}
  \end{center}
  \caption{Sensori-motor loop: The figure above shows the sensorimotor loop as
  a causal Bayesian graph. 
  \underline{Right-hand side:}
  The black circle on the left-hand side indicates the initial distribution of
  world state $W_0$, the sensor state $S_0$, the action state $A_0$ and the
  memory $M_0$ at time $t=0$. The sensor state $S_t$ depends only on the current
  world state $W_t$. The memory state $M_{t+1}$ depends on the last memory state
  $M_t$, the
  previous action $A_t$, and the current sensor state $S_{t+1}$. The world state
  $W_{t+1}$ depends on the previous state $W_t$ and on the action $A_t$.
  We do not draw a connection between the action $A_t$ and the
  memory state $M_{t+1}$, because  we clearly distinguish between inputs and
  outputs of the memory $M_t$ (which is equivalent to the controller). Any input
  is given by a sensor state $S_t$, and any output is given in form of the
  action state $A_t$. The system may not monitor its outputs $A_t$ directly, but
  through a sensor, hence the sensor state $S_{t+1}$. This is consistent with
  the figure on the left hand side, taken from
  \protect\citep{Pfeifer2007Self-Organization-Embodiment-and}}
  \label{fig:sensorimotor-loop}
\end{figure}

The general notation is that cognitive systems are situated and embodied
\citep{Brooks1991Intelligence-Without-Reason}, which means that they have a body
and live in an environment. In this understanding, the environment is everything
that surrounds and affects the system, and it is also called the system's
\emph{Umwelt} \citep{Uexkuell1957A-Stroll-Through}. We use the terminology world
$W_t$, and by that we mean the system's \emph{Umwelt} and the system's body. The
subscript $t$ denotes the state of the world at a specific instant in time. For
simplicity, we assume discrete time ($t\in\mathbb{N}$). The system does not have
direct access to $W_t$. To gain information about the world, the system requires
sensors, which generate sensor states $S_t$ (see
Fig.~\ref{fig:sensorimotor-loop}). From these sensor states, the system builds
an internal abstraction or memory $M_t$, from which it generates its actions
$A_t$. The actions affect the world, which closes the loop by generating,
together with the previous world state $W_t$, a new world state $W_{t+1}$. As
indicated by the indices, we assume that no time is required from an event that
occurs in the world $W_t$, to its response $A_t$. This is in accordance with
most mobile robots simulators, which freeze the controller while the world is
processed, and vice versa. In a more general setting, different time indices
must be chosen for every quantity, but at this point, it is sufficient to assume
instantaneousness. We will use the Greek letters $\alpha, \beta, \ldots$ to
denote generative kernels, i.e.~kernels which describe an actual underlying
mechanism or a causal relation between two quantities or states. In the causal
graphs, these kernels are represented by direct connections between the
corresponding nodes. This notation is used to distinguish generative kernels
from others, such as the conditional probability of $S_t$ given $M_{t-1}$ which
can be calculated or sampled, but which does not represent a direct causal
relation between $M_{t-1}$ and $S_t$ (see Fig.~\ref{fig:sensorimotor-loop}).
Additionally, capital letters ($A$,$B$,\ldots) denote random variables,
lower-case letters ($a$,$b$,\ldots) denote specific values that the random
variables can take, and calligraphic letters ($\mathcal{A}$,
$\mathcal{B}$,\ldots) denote sets of possible values for each random
variable.

\begin{figure}[ht]
  \begin{center}
    \includegraphics[width=16cm]{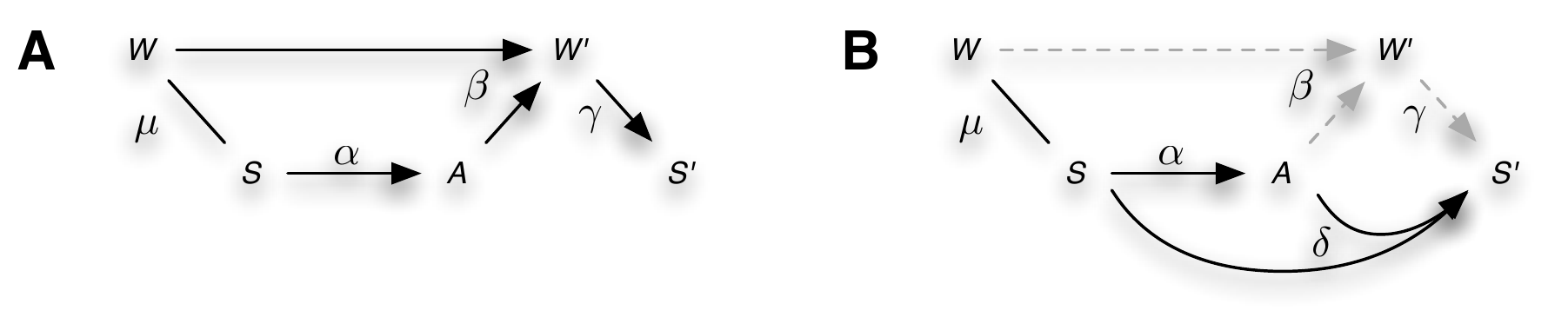}
  \end{center}
  \caption{\underline{A:} This graph shows a reduced version of
  Figure~\protect\ref{fig:sensorimotor-loop}. It shows the progression from one
  time step $t$ to $t+1$, but with different labeling to highlight the section
  of the graph which is of interest here. The random variables $A,W,S$ denote
  the present, given by some distribution $\mu$, while $W',S'$ denote the future.
  The kernel $\alpha(a|s)$ defines the policy, i.e.~which action $a$ will be
  chosen, if the sensor state $s$ is seen. Similarly, the kernels
  $\beta(w'|w,a)$ and $\gamma(s|w)$ denote the evolution of the world in
  dependence of the action and last world state ($\beta$) and the effect of the
  world on the sensor state ($\gamma$). \underline{B:} This graph shows how a
  learning rule is derived from this Bayesian network. As $\beta$ and $\gamma$
  are not available to the system, it has to build an internal world model
  $\delta(s'|a,s)$ to compensate for $\beta$ and $\gamma$.}
  \label{fig:sensorimotor-loop2}
\end{figure}

In a first step, we reduce the complexity by looking at reactive controllers,
i.e.~we omit the explicit memory $M$. We call it explicit memory, to distinguish
it from the implicit memory that is given by the adaptation of the policy due to
the sensor history.
Actions $A_t$ are now generated as a result of the
current sensor values $S_t$. In this reduced Bayesian graph (see
Fig.~\ref{fig:sensorimotor-loop2}A), the notation is changed for readability.
The past is denoted by plain letters ($A,S,W$), and the future by primed letters
($S',W'$). Excluding $M$ from Figure~\ref{fig:sensorimotor-loop}, it is
qualitatively equivalent to Figure~\ref{fig:sensorimotor-loop2}A.

Before the derivation of the learning rule can be discussed, we need to give the
basic notations of entropy and mutual information. The entropy $H(X)$ of a random
variable $X$, measuring its uncertainty, is defined as:
\begin{align}
H(X) = -\sum_{x\in \mathcal{X}} p(x) \log p(x). \label{eq:entropy}
\end{align}
All calculations in this work are given with respect to the base two logarithm
$\log_2$.
The mutual information of two random variables $X$ and $Y$ is used in this paper
in the following form:
\begin{align}
I(X;Y) = H(X) - H(X|Y). \label{eq:mutual information}
\end{align}
It measures how much the knowledge of $Y$ reduces the uncertainty of $X$, and it
is symmetric, i.e.~$I(X;Y) = I(Y;X)$ \citep{Cover2006Elements-of-Information}.
The upper bounds of the entropy and the
mutual information are needed later in this work. The maximal entropy is the
entropy of a uniform distribution and is, in this case, given by
$H(X) \leq \log_2 |\mathcal{X}|$. The equation
(see eq.~\ref{eq:mutual information}) shows that the mutual information is
naturally bounded by the entropy $H(X)$.

We can now define quantities of interest within the resulting compact causal
Bayesian network. The mutual information of past and future sensor values, known
as the predictive information \citep{Bialek2001Predictability-Complexity-and},
has been shown to be the most natural complexity measure for time series
\cite<see>[for a discussion]{Bertschinger2008An-information-theoretic}. It is
also known as excess entropy
\citep{Crutchfield1989Inferring-statistical-complexity} and effective measure
complexity \citep{Grassberger1986Toward-a-quantitative}, and plays an important
role in the related work of
e.g.~\citet{Still2009Information-theoretic-approach-to} on interactive learning.

Predictive information is defined as $I(S_p;S_f)$, where $S_p = (\ldots,S_{-2},
S_{-1}, S_0)$ is the past, and $S_f = (S_{1}, S_{2}, S_{3}, \ldots)$ is the
future of all sensor states with respect to the current time step $t=0$. It can
also be understood in terms of entropies as the reduction of the uncertainty of
the future, given the past.

It is impossible to sample the entire past and future of a system in any
concrete implementation. Therefore, we use a first order approximation of the
predictive information, i.e.~the mutual information of two consecutive sensor
values $I(S_{t+1}; S_t)$, denoted by $I(S';S)$, as the quantity of interest
here. It will not come close to the actual predictive information $I(S_p;S_f)$
as any embedded system, in general, is far from being a Markov'ian system,
i.e.~a system in which the current state only depends on the previous state.
Nevertheless, the applicability of this approximation has been shown in previous
work \citep{Ay2008Predictive-information-and,Der2008Predictive-information-and}.
To increase the readability, the term predictive information is used instead of
approximated predictive information in the remainder of this work, but we always
refer to $I(S';S)$ instead of $I(S_p;S_f)$ and we use the abbreviation PI for it. The goal
of this work can now be reformulated using this terminology. We are looking for
a learning rule that maximizes the predictive information $I(S';S)$ by modifying
the policy $\alpha(a|s)$.

The calculation of the PI relies on knowledge about the progression of the
world, i.e.~knowledge about the kernels $\beta(w'|w,a)$ and $\gamma(s'|w')$.
These kernels are not accessible to the system, as a
cognitive system can only rely on information that is intrinsically available.
To solve this problem, we introduce an intrinsic world model $\delta(s'|a,s)$
to replace $\beta$ and $\gamma$ (see Fig.~\ref{fig:sensorimotor-loop2}B).
This replacement is valid, as the joint probability distribution $p(s',s)$ is sufficient to
calculate $I(S';S)$:
\begin{align}
  I(S';S) & = \sum_{s',s\in\mathcal{S}} p(s',s)
  \log_2\frac{p(s',s)}{p(s')p(s)},\label{eq:mi 2} & p(s) & =
  \sum_{s'\in\mathcal{S}} p(s',s), &  p(s') =
  \sum_{s\in\mathcal{S}} p(s',s), 
\end{align}
and we can deduce from Figure~\ref{fig:sensorimotor-loop2}B that:
\begin{align}
p(s',s) &= \sum_{a \in \mathcal{A}} p(s,a,s') = \sum_{a\in\mathcal{A}} p(s) \alpha(a|s) \delta(s'|s,a)
 \label{eq:p(s'|s)}.
\end{align}
This shows that the predictive information $I(S';S)$ can be calculated without
$\beta$ and $\gamma$ if the sensor distribution $p(s)$, the policy $\alpha(a|s)$
and the intrinsic world model $\delta(s'|s,a)$ are known. This intrinsically
calculated PI converges to the actual PI (for a stationary process) as the sampled intrinsic world model
$\delta(s'|s,a)$ converges against the actual world
model. As will be shown later,
this is the case in the experiments presented in this work (see
Sec.~\ref{sec:experiments}).

Now, the natural gradient \citep{Amari1998Natural-Gradient-Works} of the
predictive information can be calculated with respect to the policy
$\alpha(a|s)$ (see app.~\ref{sec:appendix derivation} for details). In a first
step, we represent the distributions $p(s)$, $\alpha(a|s)$, and $\delta(s'|s,a)$
as matrices. This explicitly means, that we do not restrict the possible
probability distributions and models. Next, the update equations for the sensor
distribution, the world model and the policy are given.

The sensor distribution is simply sampled over time and it is given by:
\begin{align}
  p^{(0)}(s) & :=  \frac{1}{|\mathcal{S}|}\nonumber\\
  p^{(n)}(s) & :=  \left\{\begin{array}{cl}
  \displaystyle\frac{n}{n+1}p^{(n-1)}(s)+\frac{1}{n+1} & \text{if } S_{n+1} = s\\[5ex]
  \displaystyle\frac{n}{n+1}p^{(n-1)}(s) & \text{if } S_{n+1} \not= s
  \end{array}\right. \label{eq:mil sensor distribution}
\end{align}
The update rule for the world model is very similar to the rule for the updating of the
sensor distribution and reads:
\begin{align}
  \delta^{(0)}(s'|s,{a})                  & :=  \frac{1}{|S|}\nonumber\\
  \delta^{\left(n_{a}^s\right)}(s'|s,{a}) & :=  \left\{\begin{array}{ll}
  \displaystyle\frac{n_{a}^s}{n_{a}^s+1}\delta^{(n_{a}^s-1)}(s'|s,{a})+\frac{1}{n_{a}^s+1} 
                  & {{\text{if } {S_{n_{a}^s+1} = s',\, S_n=s,\,A_{n_{a}^s+1}={a}}}}\\[3ex]
  \displaystyle\frac{n_{a}^s}{n_{a}^s+1}\delta^{(n_{a}^s-1)}(s'|s,a) 
                  & {\text{if } S_{n_{a}^s+1} \not= s',\, S_n=s,\,A_{n_{a}^s+1}={a}}\\[3ex]
  \delta^{(n_{a}^s-1)}(s'|s,{a})
  & {\text{if } S_{n_{a}^s}\not=s \text{ or } \,A_{n_{_{s,{a}}}+1}\not={a}}
  \end{array}\right. \label{eq:mil world model}
\end{align}
What is important to note here is that there is a counter $n_{a}^s = 1,2,\ldots$ for every pairing of
$(s,a)$ to assure that the learning rate for each row of the world model matrix,
i.e.~each pairing of $(s,a)$, decays according to the
number of samples in that row, and not faster. 
The update rule for the policy (see below) seems complex due to the fact that
there are no a priori assumptions about the probability distribution. Using
the natural gradient on the predictive information, we get:
\begin{align}
  \alpha^{(0)}({a}|s) & :=  \frac{1}{|S|} \nonumber\\
  \alpha^{(n)}({a}|s) & = \alpha^{(n-1)}({a}|s) +
  \frac{1}{n+1} \alpha^{(n)}({a}|s)
  \left(F(s) - \sum_{a} \alpha^{(n-1)}(a|s) F(s)\right)   \label{eq:mil policy}\\
  F(s) & := p^{(n)}(s)\sum_{s'}\delta^{(n)}(s'|s,{a})
  \log_2\frac{\sum_{{a}}\alpha^{(n-1)}({a}|s)\delta^{(n)}(s'|s,{a})}
    {\sum_{s''}p^{(n)}(s'') \sum_a \alpha^{(n-1)}(a | s'') \, \delta^{(n)}(s' | s'',a)}
    \nonumber
\end{align}
Note that our learning rate $\frac{1}{n+1}$ satisfies the standard condition for
corresponding rates in stochastic approximation theory, that is
$\sum_{n=1}^{\infty} a_n^2 < \infty$, 
$\sum_{n=1}^{\infty} a_n = \infty$
\citep{Benveniste1990Adaptive-algorithms-and}. These assumptions are usually
made in order
to ensure convergence. However,
a possible point of criticism here is that the learning and update factors
$\frac{1}{n_a^s + 1}$ and $\frac{n_s^a}{n_a^s + 1}$ are not well chosen, as they
lead to strong changes during the first iterations and because they may converge
too fast towards zero. Other adaptation factors were evaluated, and are
discussed later in this paper (see Sec.~\ref{sec:results}).

Now that the learning rule is defined, the next step is to evaluate its effect
on the PI and the behavior of a system in the sensorimotor loop. This is discussed in the
next section.

\section{Experiments}
\label{sec:experiments}
The experiments chosen for presentation here are inspired
by the previous work of \citet{Ay2008Predictive-information-and} and
\citet{Der2008Predictive-information-and}. In the former publication,
individually controlled, simple two-wheeled differential drive robots were
physically coupled to form a chain which operates in a bounded, featureless
environment, while in the latter, a single robot was placed in a bounded
environment with cubical obstacles.

In both publications, the predictive information is used in the context of the
sensorimotor loop. In \citep{Der2008Predictive-information-and} a robot chain
of length five is equipped with simple parametrized controllers. For each
parameter setting, a series of experiments were performed, and the predictive
information was then calculated based on the recorded time series. It is shown,
that the predictive information is high for parameter settings which also show a
high coordination among the robots. The coordination among the robots is
measured indirectly by the entropy over the probability distribution of the
position of the center robot in the bounded environment. This form of indirect
measure of the coordination is also used in this work, but in contrast, this
work will also analyze the effect of modifications to the robot chain length on
the predictive information and the coordination among the robots.

The work by \citet{Ay2008Predictive-information-and} presents a learning rule
that maximizes the predictive information under certain constraints. One of the
constraints is that the world model $\delta(s'|s,a)$ is 
modeled by a deterministic function with additive Gaussian noise. The result
is a learning rule that is equivalent to the Homeokinetic principle by
\citet{Der2001Self-Organized-Acquisition-of,Der2002True-autonomy-from}.

In summary, this work differs from the previous work in two aspects. First,
a learning rule, which is unrestricted with respect to the underlying model and free of
assumptions on the world $W_t$ is derived and evaluated. Second, the contribution
of the chain length to the PI is also analyzed.

\subsection{Experimental Setup}
The remainder of this section presents the setup of the simulator, the robot,
the controller, and finally of the environment, before the following section
discusses the results of the experiments.

\paragraph{Simulator:} All experiments were conducted purely in simulation for
the sake of simplicity, speed and analysis. It is faster to setup and conduct
experiments in simulation, with respect to experiments with real robots, as well
as to generate and record the data necessary
for analysis. Current simulators, such as YARS
\citep{Zahedi2008YARS:-A-Physical}, which was chosen in this work, are shown to
be realistic enough to simulate the
relevant physical properties of mobile robots, and designed such that
experimental runs can be automated, run at faster than real-time speed, and require 
minimum effort to setup.
\paragraph{Robots:} In the following paragraphs, we will talk of a chain of
identical robots (or chain for short), and thereby mean any number of physically
coupled robots including a single robot. A robot in this chain is derived from
the Khepera I robot \citep{Mondada1993Mobile-Robot-Miniaturization:}, which is a
two-wheeled differential drive robot with a circular body (see
Fig.~\ref{fig:experimental setup}).

\begin{figure}[h]
\begin{center}
  \includegraphics[width=16cm]{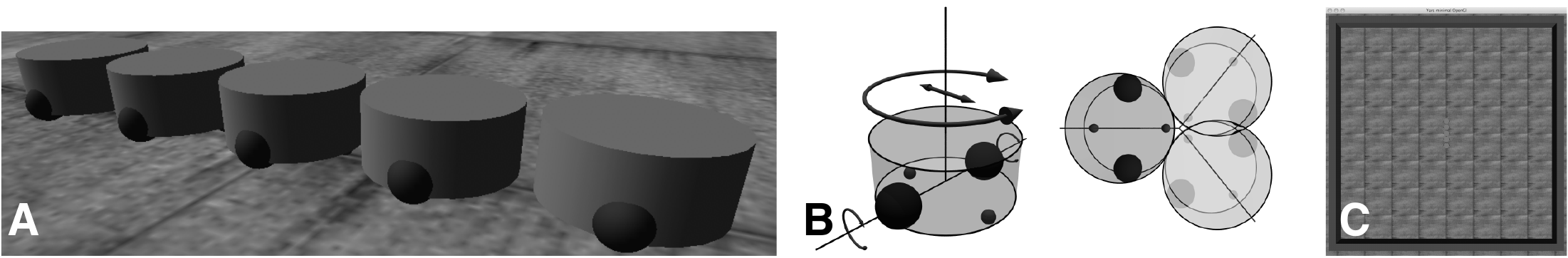}
\end{center}
\caption{Experimental setup: Figure (B) shows a sketch of the
  two-wheeled differential drive robot and the connection between neighboring
  robots, Figure (A) a chain of five robots and Figure (C) and the bounded,
  featureless environment used in the YARS simulator.}\label{fig:experimental
  setup}
\end{figure}

In the experiments presented here, the only inputs and outputs of the robot are
its desired wheel velocity ($A_t$), and the current actual wheel velocity
($S_t$). Both quantities are mapped linearly to the interval of $[-1, 1]$, where
$-1$ refers to the maximal negative speed (backwards motion), and $+1$ to the
maximal positive speed (forward motion). No noise is artificially added to the motors
or sensors.

The robots are connected by a limited hinge joint with a maximal deviation of
$\pm0.9$ rad ($\approx 100$ degree), thereby avoiding intersection of
neighboring robots (see Fig.~\ref{fig:experimental setup}).

\begin{figure}[ht]
\begin{center}
  \includegraphics[width=8cm]{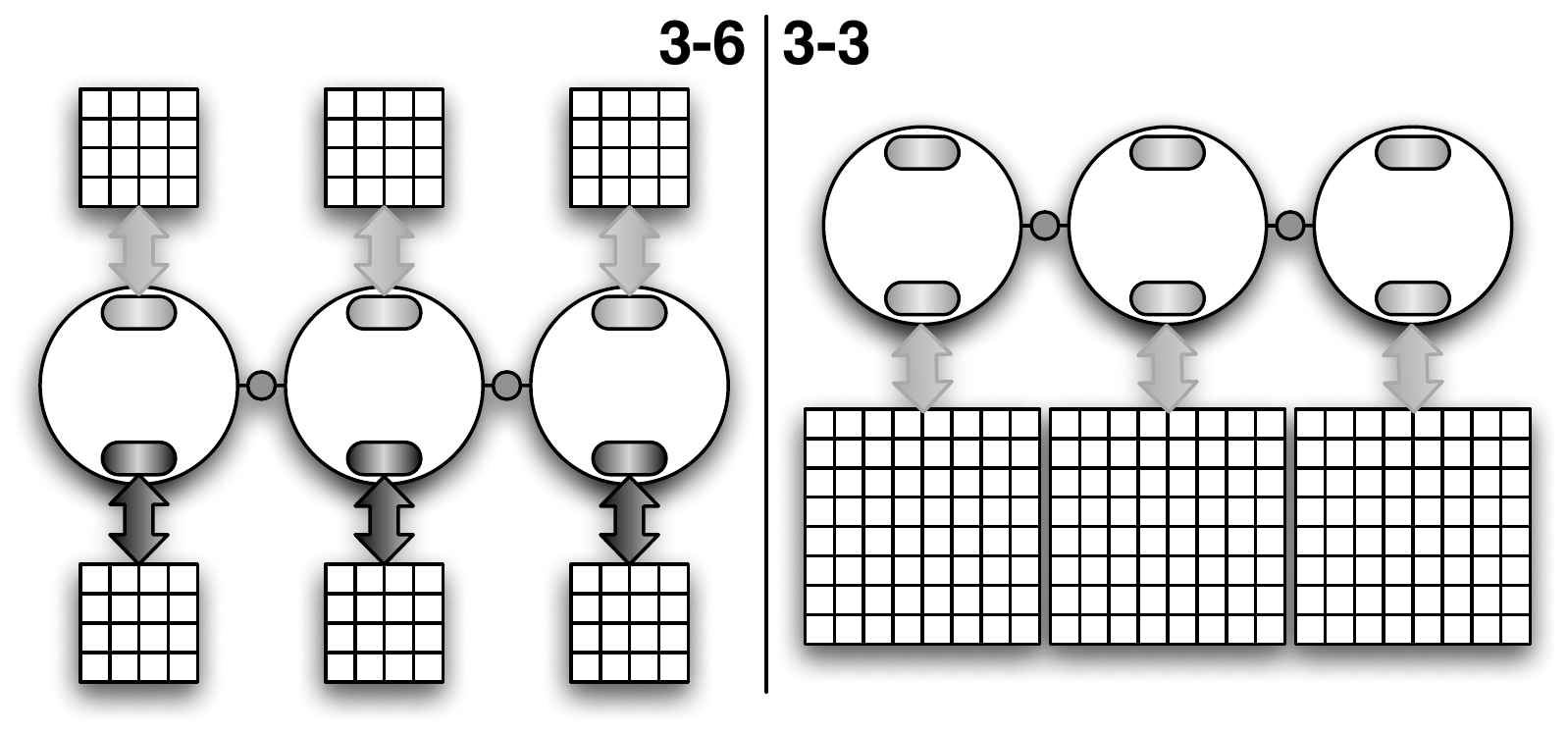}
\end{center}
\caption{Controller Setup. The labels \conf{3}{3} and \conf{3}{6} refer to
three robots with three controllers, and three robots with six controllers,
respectively. \underline{Left-hand side:} Split controller setup.
Each wheel of each robot has an individual controller, i.e.~policy matrix
$\alpha(a|s)$. \underline{Right-hand side:} Combined controller setup.
Each robot has an individual controller, i.e.~policy matrix
$\alpha(a|s)$. If the policy matrix $\alpha$ has the size $n\times n$ in the
split case, the policy matrix in the combined case has the size $n^2 \times
n^2$. For the sake of conciseness, the latter is indicated by a matrix of size
$2n \times 2n$, not $n^2 \times n^2$.}
\label{fig:controller setup}
\end{figure}

Three different kinds of experiments are presented in the following section, single robot,
three-, and five-segment chains. Chains of two and four robots were also tested, but not
chosen for presentation here, as their analysis did not provide additional
insights.
\paragraph{Controller:} Inspired by
\citet{Ay2008Predictive-information-and} and \citet{Der2008Predictive-information-and}, each
robot is controlled locally, i.e.~there is no global control which has access to
every wheel of every segment. For the local control, two control paradigms are evaluated;
\emph{combined} and \emph{split} control (see Fig.~\ref{fig:controller setup}). The former refers to a single controller for both
wheels, while in the latter case each wheel has its own controller. There is no
communication between the controllers. Any interaction occurs
solely through the world $W_t$, and hence, through the sensor states $S_t$,
which are in this case only the current actual wheel velocity.
The controllers run at 10Hz. Every experiment was setup to run for $10^6$
controller updates, resulting in an overall time for each run of approximately
27.5 simulated hours.

In the work presented here, due to the discretisation and matrix representation,
pre- and post-processing in the form of binning is required. We chose four
equally distributed bins in the interval $[-1,1]$ for the input and output
spaces. Different numbers of bins, from 3 to 30, were evaluated. While three bins
were dismissed because the result was too close to defining three disjunct
actions (forward, stop, backwards), a higher number of bins ($>8$) resulted in
less coordination among the robots (compared to four bins).

For the remainder of this work, the notation \conf{$r$}{$c$} is used, where
$r\in\{1,3,5\}$ defines the number of robots, and $c\in\{r,2r\}$ gives the
number of controllers. Therefore, the label 1-1 refers to a single robot with a
single controller for both wheels, and, at the other end, \conf{5}{10}
refers to a chain of five robots, with ten controllers, i.e. one for each wheel
(see Fig.~\ref{fig:controller setup}).

\paragraph{Environment:} The environment is a bounded, but otherwise
featureless, chosen large enough for the chains to be able to learn a
coordinated behavior. Each of the robots has a size of $10cm$ in diameter.
The environment's size is eight by eight meters. Every chain was started
with its center robot in the center of the environment, and with the same
initial heading.

\section{Results} 
\label{sec:results}
This section discusses the results from the six experiments presented above. The
presentation of the results is given in the following steps. First, it is
analyzed if the PI was increased over time for all six configurations, and if
so, how close it gets to the theoretical upper bounds. In the next step, the
increases of the PI over time are related to modifications of the behavior,
answering the question if the maximization of the PI leads to qualitative
changes on the behavior. The third step is to quantify the behaviors
for comparison. From these findings, a seventh experiment is derived and
analyzed, in which a combined controller of the \conf{1}{1} configuration is
initiated with the two optimal split controllers from the \conf{1}{2}
configuration.

\subsection{Maximizing the Predictive Information}
\begin{figure}[h]
  \begin{center}
    \includegraphics[width=16cm]{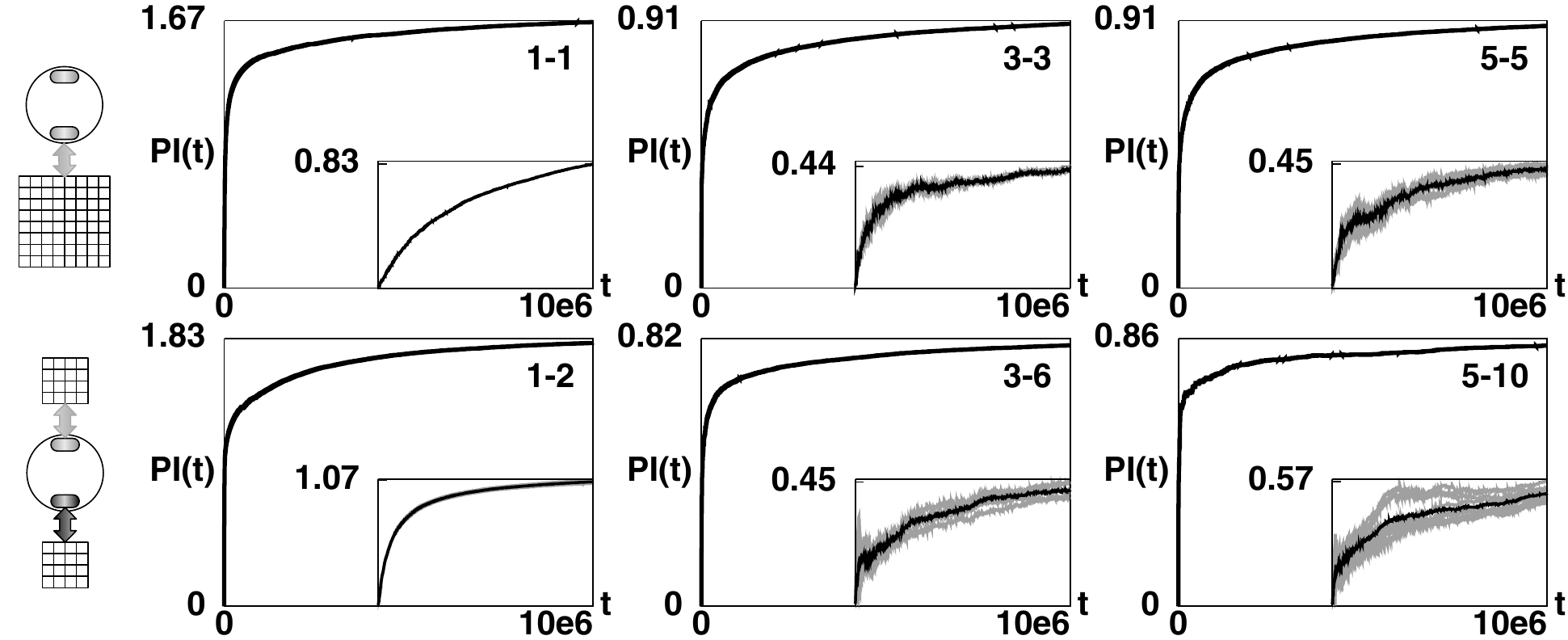}
  \end{center}
  \caption{Average-PI plots for each of the six experiments. The ordering from
  upper left to lower right is, \conf{1}{1}, \conf{3}{3}, \conf{5}{5},
  \conf{1}{2}, \conf{3}{6}, \conf{5}{10}. The evolution of the average PI is
  shown in black in each plot. The embedded
  small plots show the progress for the first five minutes of each run,
  including the PI for each controller, plotted in gray.}
  \label{fig:mi plots}
\end{figure}

The first step is to analyze and compare the development and the maximally
achieved values
of the predictive information for the six settings.
The Figure \ref{fig:mi plots} shows the progression of the predictive
information over the entire time, with embedded plots for the initial learning
phase. The tables (Tab.~\ref{tab:mi comparison}) show the
final PI values for the six experiments, and PI values
calculated as an external observer. How the latter is calculated will be
explained later in this section.

\begin{table}
  \begin{center}
    \begin{minipage}[c][4cm][t]{6cm}
      \begin{tabular}{l||c|c|c}
        \multicolumn{4}{c}{Final intrinsic PI}\\
        c / r & 1              & 3              & 5\\
        \hline\hline
        combined & 1.66  & 0.90 & 0.89 \\
                 &  42\% & 23\% & 23\% \\
         \hline
        split    &  1.80 &  0.80 & 0.84 \\
                 &  92\% &  42\% & 42\% 
      \end{tabular}
    \end{minipage}
    \begin{minipage}[c][4cm][t]{6cm}
      \begin{tabular}{l||c|c|c}
        \multicolumn{4}{c}{PI on recorded data}\\
        c / r & 1              & 3              & 5\\
        \hline\hline
        combined & 2.60 & 1.59 & 1.70 \\
                 & 27\% & 16\% & 17\% \\
        \hline
        split    & 4.01 & 2.25 & 2.39 \\
                 & 41\% & 23\% & 24\%\\[0.2ex]
         \multicolumn{4}{r}{(upper bound $\log_2(30^2)\approx 9.81)$}
      \end{tabular}
    \end{minipage}
  \end{center}
  \caption{Comparison of intrinsically calculated PI (left-hand side) and PI calculated a
  posteriori on the recoded data per robot (right-hand side). The ordering is
  roughly kept, but more differentiated, which is also the result of the higher
  binning (4 vs. 30 bins).}
  \label{tab:mi comparison}
\end{table}

The first result from the Figure \ref{fig:mi plots} is that the learning rule as
defined in the equations (\ref{eq:mil sensor distribution})-(\ref{eq:mil
policy}) successfully increases the PI in all six configurations.
To be able to understand how well the PI
is maximized, the values are compared to their theoretical upper bounds, which
can be calculated from the chosen binning. In the split controller case, the
theoretical upper bound is given by $\log_24=2$, while the upper
bound for the combined case is given by $\log_216=4$. The comparison of the
achieved PI values with their upper bounds is shown in Table 
\ref{tab:mi comparison}. The configuration \conf{1}{2} almost
achieves is maximum, while the configurations \conf{1}{1}, \conf{3}{6},
\conf{5}{10} are equally successful with about 42\% of their maximum. The
configurations \conf{3}{3} and \conf{5}{5} have the smallest relative PI with
about 23\%. 

The results are not well comparable across controller paradigms, as two
processes, one with a single channel (split control) and the other with two
channels (combined control) are compared. A
better way is to compare the PI per robot, and
hence, the PI is calculated additionally on the recoded
sensor data. This is done in the
following way. For each time step, the actual sensor values $S_l(t), S_r(t) \in
[-1,1]$ were recorded and binned into thirty equally distributed bins for each
wheel. From this data, the mutual information $I(\{S_l(t), S_r(t)\}; \{S_l(t+1),
S_r(t+1)\})$ is calculated over the last $5\cdot10^5$ points in the data set and
the results are shown in Table \ref{tab:mi comparison}.
The results allow a direct comparison of the configurations and the
values and
show that the chains of length three and five maximize comparably well, although
the split controller configurations show slightly higher values. For the single
robot configuration, both are significantly higher compared to the multi-robot
chains, and again, the split controller outperforms the combined controller with
respect to the PI achievement.

At this point, the conclusion is, that the single robots configurations succeed
better in maximizing the predictive information compared to longer chains, and
that in general split controller outperform the combined counterpart. 
The next step is to relate these findings with the behaviors of the systems.

\subsection{Comparing behaviors}
\label{sec:comparing behaviours}
Before the behaviors are analyzed and related to the results of the previous
section, we briefly repeat what the predictive information measures. The
predictive information ($I(S';S) = H(S) - H(S'|S)$) is high, if the sensor entropy $H(S)$ is high, and if the
uncertainty of the future given the past $H(S'|S)$ is low. Applied to the
chains, we expect a high predictive information if the controllers have high
wheel velocity variance ($H(S)$), but at the same time low variance in the changes
of the wheel velocity ($H(S'|S)$). 
This means that each configuration should try to sense every wheel velocity
with almost same probability, and at the same time be as deterministic as
possible. 

\begin{figure}[ht]
  \begin{center}
    \includegraphics[width=8cm]{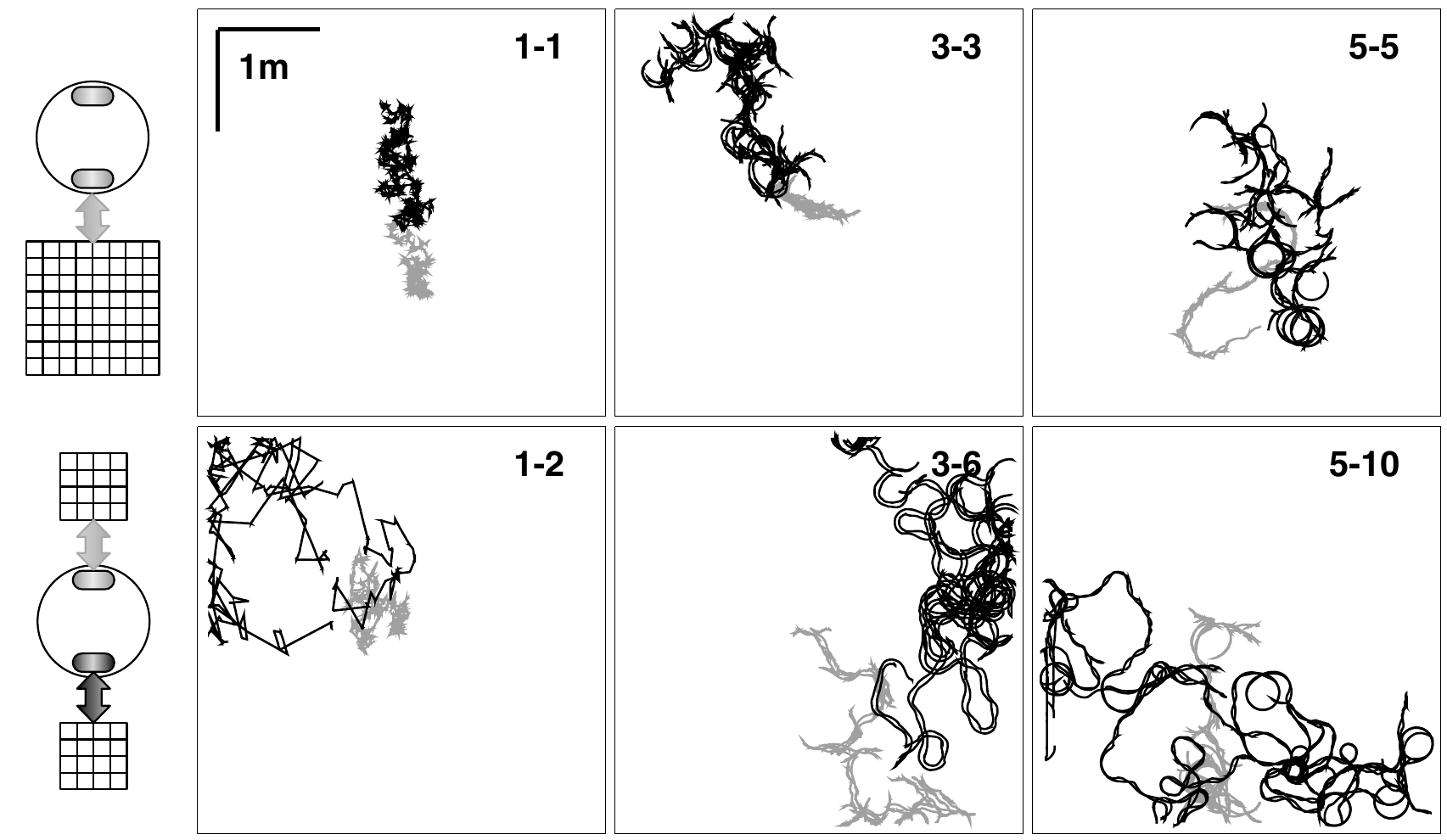}
  \end{center}
  \caption{Trajectories: These six plots show the trajectories of the six systems
  for the first ten minutes (Grey) and the last 100 minutes (black). From upper left to
  lower right: \conf{1}{1}, \conf{3}{3}, \conf{5}{5}, \conf{1}{2}, \conf{3}{6},
  \conf{5}{10}. The plots show that the length of the trails with consecutive
  movement increases with the number of robots, and with split controllers.}
  \label{fig:trajectory}
\end{figure}

The trajectories cannot be visualized entirely, as the resulting plots would not show
distinguishable trails. Therefore, the Figure \ref{fig:trajectory}
shows the first ten minutes in Grey, and the last 100 minutes
in black in the foreground. The Grey trajectory shows the behavior during the
initial learning phase, while the black trajectory shows the converged behavior.

The single robot with split controller (\conf{1}{2})
shows straight and rotational movements. The chains show wavy lines and
alternating headings. 
To better differentiate the behaviors, two quantification methods are used,
which are both explained in the following paragraphs.

\begin{figure}[ht]
  \begin{center}
    \includegraphics[width=16cm]{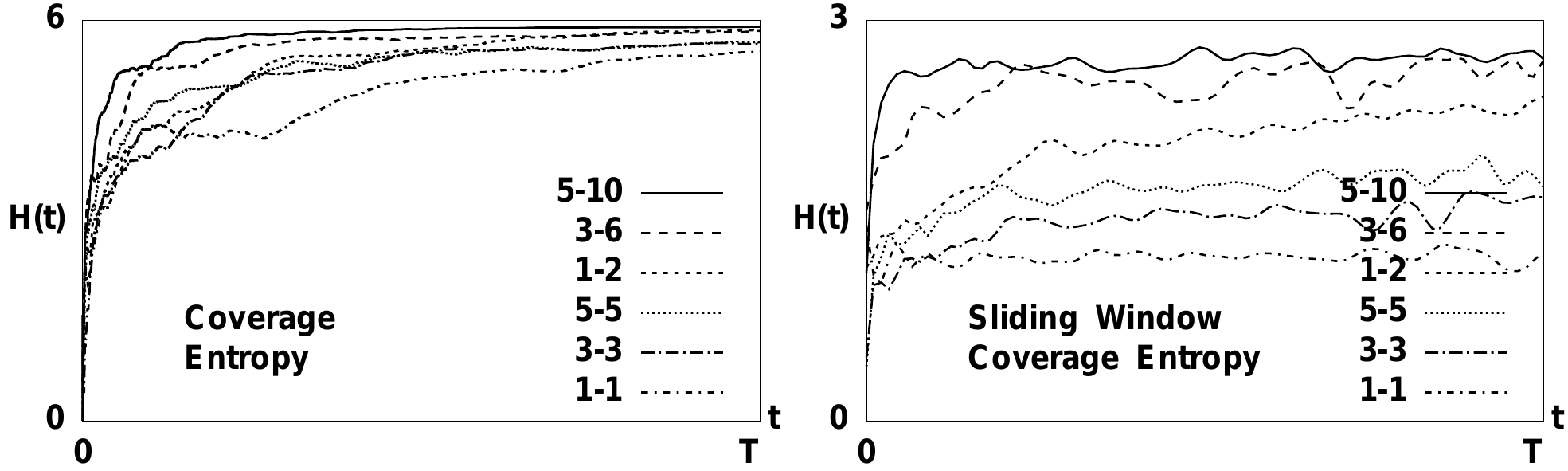}
  \end{center}
  \caption{Coverage Entropy: \underline{Left-hand side:} The overall coverage
    entropy (see text for detailed explanation). The plot shows that every
    configuration finally covers the entire space (in $T=10^6$ time steps). The
    ranking of the label corresponds to the speed with which the coverage is
    achieved.
    \underline{Right-hand side:} Coverage entropy for a sliding window of ten
    seconds, plotted as a B\'ezier curve ($T\approx10^6$ time step). This plot shows
    that \conf{5}{10} is the fastest to achieve a good exploration behavior,
    followed by \conf{3}{6}. The configuration \conf{1}{2} requires much more
    time, but eventually shows comparable exploration behavior. The robots with
    combined control of the wheels are all outperformed by those with split
    control.}
  \label{fig:coverage entropy}
\end{figure}

The first quantification method is the coverage entropy used by
\citet{Der2008Predictive-information-and}. The bounded space is divided into 400
($20\times20$) patches of equal size. At every time step, the position of the
center robot is measured, and the counter for the corresponding patch is
increased. This gives a visit frequency for each patch, from which the 
coverage entropy is calculated (see
Fig.~\ref{fig:coverage entropy}, left-hand side). It is plotted over time, as it
then shows how fast a system covers the entire space.

The second method is designed to show how the behaviors vary over time.
In this second case, the coverage entropy is calculated on a sliding window. For the
Figure \ref{fig:coverage entropy} (right-hand side), a sliding window of width
$10^3$ was chosen. The resulting $10^3$ values are used as control
points for a B\'ezier curve\footnote{The plots were generated with gnuplot
\protect\shortcite{Williams2009gnuplot-4.2.6} and its internal B\'ezier implementation.}. 

The results of both methods (see Fig.~\ref{fig:coverage entropy})
allow the following conclusions:
\begin{enumerate}
  \item All configurations explore the entire area
    (see Fig.~\ref{fig:coverage entropy}, left-hand side), but require different
    time.
  \item Longer consecutive trails relate to higher average sliding window
    coverage entropy (compare Fig.~\ref{fig:trajectory} with the right-hand side
    of Fig.~\ref{fig:coverage entropy}).
  \item The configurations which show longer consecutive trails are those, which
    reach higher coverage entropy sooner. 
\end{enumerate}
As stated earlier, movements only occur for chains with length larger than one
if the majority of the segments moves in one direction.
Therefore, the sliding
window coverage entropy allows us to indirectly measure the cooperation of the segments.
 We therefore see higher cooperation among the segments of the
split configuration, when compared to their combined controller counterparts.
This will be discussed at the end of this paper (see Sec.~\ref{sec:discussion}).

Obviously, the measures do not relate to cooperation for the single robot
configurations, but the measures also show here that higher PI relates to higher
coverage entropy and higher sliding window coverage entropy, for the split
controller paradigm.

Configuration \conf{1}{2} is the only one to achieve almost maximal PI.
Therefore, its strategy is chosen for analysis in the next section.
In addition, the behavior of the configuration \conf{3}{6} is analyzed, as it
reveals why longer chains result in longer trails. Furthermore, it shows that
the solution for the chains with more than one robot is not binning-specific.

\subsection{Behavior Analysis:} 
\begin{figure}[ht]
  \begin{center}
    \includegraphics[width=16cm]{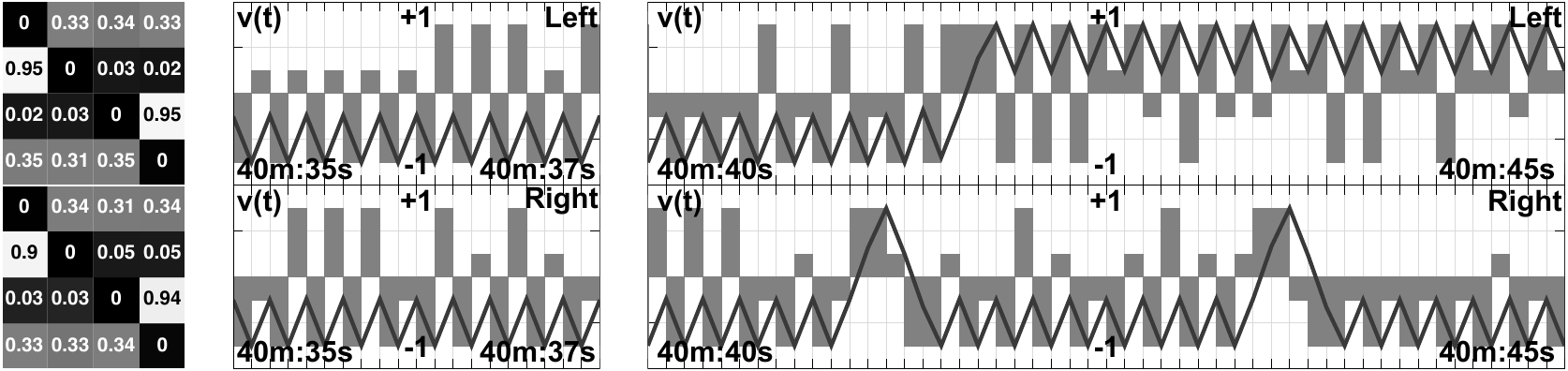}
  \end{center}
  \caption{\conf{1}{2}: \emph{Left:} Policies for left (upper) and right (lower)
  wheel. \emph{Center:} Transient plot for a sequence of straight movement.
  \emph{Right:} Transient plot for a sequence of rotations.}
  \label{fig:1S}
\end{figure}

\begin{figure}[ht]
  \begin{center}
    \includegraphics[width=16cm]{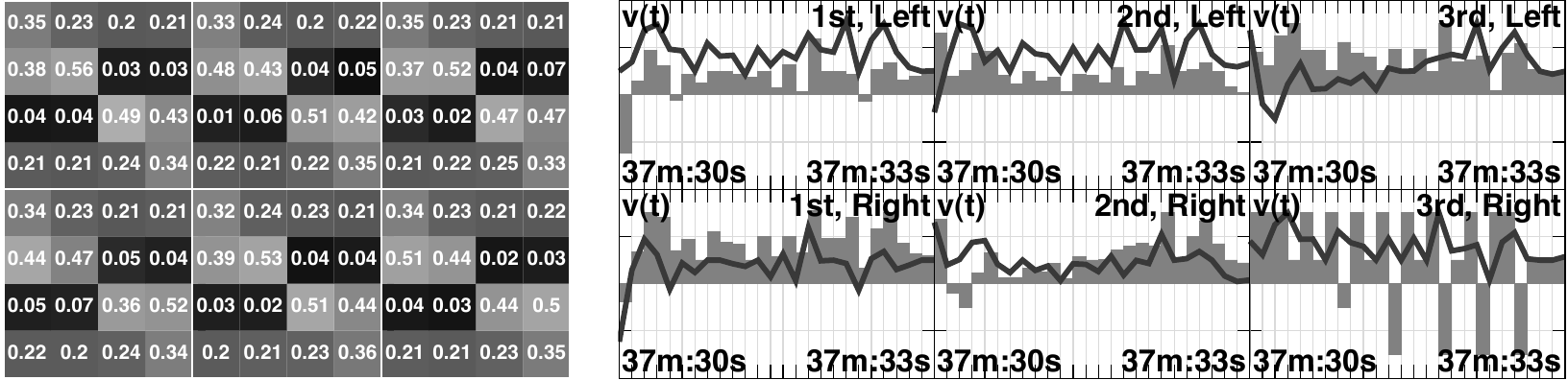}
  \end{center}
  \caption{Left-hand side: The \conf{3}{6} policy. Right-hand side: Transient
  plot.}
  \label{fig:3S}
\end{figure}

For the analysis, each configuration was fixed
after learning, and then run for one simulated hour ($3.6\cdot10^4$ iterations)
during which the controller output $A$ and the sensory input $S$ was recorded.
Figures \ref{fig:1S} to \ref{fig:3S} are taken from these recordings and show
the actions as bars, and the sensor states as lines. 

\begin{figure}[ht]
  \begin{center}
    \includegraphics[width=8cm]{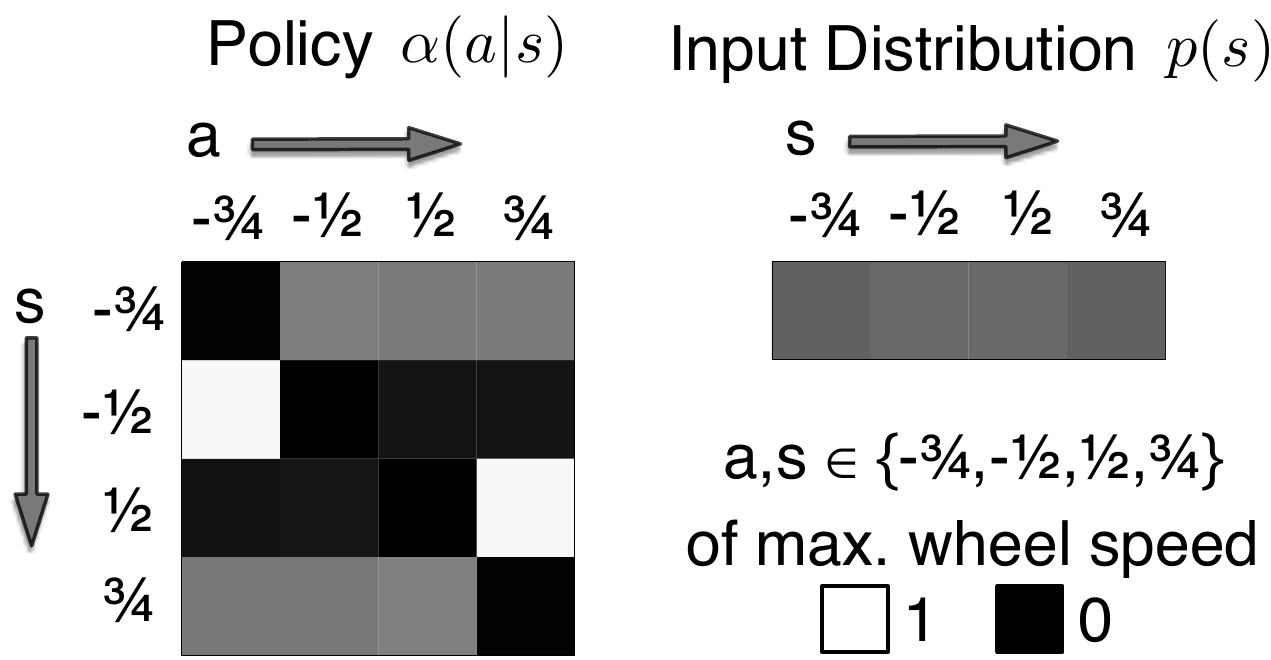}
  \end{center}
  \caption{Policy and sensor distributions of the configuration \conf{1}{2}. The
  matrices of the left wheel are shown, as the right wheel matrices do not show
  a significant difference. The matrix on the left-hand side is the policy
  $\alpha(a|s)$. The columns represent the
  action, starting from left to right, i.e.~the action representing high
  backward motion $\left( -\frac34 \right)$ is the left most column, while the action
  representing high forward motion $\left( \frac34 \right)$ is the right most column.
  Similarly, the sensor states are represented by the rows, where the top most
  row represents full backward motion and the lowest row, full forward motion.
  The numbers represent the bin centers. The vector on the right-hand side
  is the sensor or input distribution $p(s)$.} 
\label{fig:1-2 matrices}
\end{figure}

The following naming convention is used in the paragraphs ahead. The bins are
named with respect to their center, i.e.~$-\frac34$, $-\frac12$, $\frac12$,
$\frac34$. These names relate to the maximal positive ($-1$) and negative
($+1$) wheel velocities. The policy and sensor distribution configuration is
shown exemplarily for the \conf{1}{2} configuration in Figure \ref{fig:1-2 matrices}.

\paragraph{Configuration \conf{1}{2}:}
The policy and transient plots (see Fig.~\ref{fig:1S}) reveal
how the maximal PI is achieved.
The transient plot displaying straight movement (see Fig.~\ref{fig:1S}, center)
shows that the wheel velocities oscillate between $-\frac12$ and $-\frac34$.
This oscillation is stable due to the physical properties of the system for the
following reason. The sensor state $S=-\frac34$ results in an action
$A\in\{-\frac12, \frac12, \frac34\}$. Due to the inertia of the system and the
controller frequency, any selected action $A\in\{-\frac12, \frac12, \frac34\}$
leads to a sensor state $S=-\frac12$, as the desired wheel velocity cannot be
reached instantaneously. As a consequence, the action $A=-\frac34$ is chosen with a
probability of $p(A=-\frac34 | S=-\frac12) = 0.95$, leading to the
observable oscillation during the translational movement of the robot (see
Fig.~\ref{fig:1S}, center). With a remaining probability of
$p(A\not=-\frac34|S=-\frac12)=0.05$, a change of the direction of the wheel
velocity occurs, which leads either to a rotation of the system, or inversion of
the translational behavior (see Fig.~\ref{fig:1S}, right-hand side). As a
result, the sensor entropy $H(S)$ is high (compare with Fig.~\ref{fig:1-2
matrices}), but at the same time the conditional
entropy $H(S'|S)$ is low, leading to the observed high PI.

\paragraph{Configuration \conf{3}{6}} The transient plot (see Fig.~\ref{fig:3S})
clearly shows a difference to the configuration \conf{1}{2}, as the wheel velocity of
one wheel is no longer only influenced by its controller, but also by the
actions of the other controllers. To understand the strategy, the policy for
$S=-\frac34$ must be taken into account ($S=\frac34$ is analogous). With a
probability of $p(A\in\{-\frac34, -\frac12\} | S=-\frac34) \approx 0.6$, the current
direction of the wheel rotation is maintained (see Fig.~\ref{fig:3S}, left-hand side). As at least two
robots, i.e.~four related controllers must move into the same direction, for the
entire system to progress, the probability of a switch in the direction is
approximately $p\approx0.4^4$. If only one controller switches, the sensor state
remains (as discussed above), i.e.~the direction of the system is unchanged. This explains how the
robots coordinate, and why the configuration \conf{5}{10} shows longer consecutive
trails with very similar policies (not shown).

These analysis also show that the solutions are not specific to the selected
four bins configuration. In an incremental way, one can construct policies of
higher dimension by splitting of the values of the corresponding cells. An
example is a policy with eight bins, which can be constructed from the four bin
policies by the following mapping
$p_{\{8\,\, bins\}}(a_i|s_j) = \frac14p_{\{4\,\,bins\}}(a_{\lfloor i /
2\rfloor}|s_{\lfloor j/2\rfloor})$. This indicates a possibility to
incrementally increase the dimension of the policies based on adapted lower
dimension solution. A different form of incremental optimization in this
context, in which combined controllers can be constructed from split solutions,
is discussed in the next section.

\subsection{Incremental Optimization}
In the results presented above, the \conf{1}{1} configuration was least successful in
maximizing its predictive information. In general, the configuration \conf{1}{1}
should be at least as successful as the \conf{1}{2} configuration, because the
combined controller has the same capabilities that the system consisting
of two split controllers has, if not more. Hence, there are two
possible reasons why the \conf{1}{1} configuration is less successful. First, it
(repeatedly) reaches a sub-optimal solution, and second, the learning and update
rate $\frac{1}{n+1}$ has converged faster towards zero than the behavior
towards the optimal solution. To exclude the latter possibility, different types of
learning rates (bounded below, constant during an initial learning phase, both,
and constant) were evaluated, and the
experiments clearly showed that the chosen learning rate is not the reason for
the sub-optimal solution.
To test the former hypothesis, a combined controller was generated from
two split controllers, including the initial sensor distribution ($p(s)$) and
world model ($\delta(s'|s,a)$). The combined policy is generated using the
products of the two split policies in the following manner, where the superscripts $l,r,c$ refer
to split left, split right and combined controller:
\begin{align*}
\alpha^l(a^l|s^l) & = (\alpha^l_{s,a}) & a & = 0,1,\ldots, |\mathcal{A}|-1 
                                       & s & = 0,1,\ldots, |\mathcal{S}|-1 \nonumber\\
\alpha^r(a^r|s^r) & = (\alpha^r_{s,a}) \nonumber\\
\alpha^c(a^c|s^c) & = (\alpha^c_{s^c,a^c}) & a^c & = 0,1,\ldots, |\mathcal{A}|^2-1 
                                           & s^c & = 0,1,\ldots, |\mathcal{S}|^2-1\nonumber\\
\alpha^c_{s^c,a^c} & = \alpha^l_{s_c^l, a_c^l} \cdot \alpha^r_{s_c^r, a_c^r} 
                   & s_c^l & = s^c \mod |\mathcal{S}|, \quad
s_c^r = \left\lfloor\frac{s^c}{|\mathcal{S}|} \right\rfloor
                   & a_c^l & = a^c \mod |\mathcal{A}|, \quad 
                    a_c^r = \left\lfloor\frac{a^c}{|\mathcal{A}|} \right\rfloor.
\end{align*}
The equations above implement cascaded loops, such that the indices for the
combined controller $(s^c, a^c) = (s^l_c, a^l_c, s^r_c, a^r_c)$ is given by the
sequence (shown exemplarily for some elements):
$(0, 0) = (0, 0, 0, 0)$, $(0, 1) = (0, 1, 0, 0)$ $(0, 2) = (0, 2, 0, 0)$,
\ldots,
$(2, 10) = (2, 2, 0, 2)$, $(2, 11) = (2, 3, 0, 2)$, $(2, 12) = (2, 0, 0, 3)$,
$(2, 13) = (2, 1, 0, 3)$, $(2, 14) = (2, 2, 0, 3)$, $(2, 15) = (2, 3, 0, 3)$
\ldots

\begin{figure}[ht]
\begin{center}
  \includegraphics[width=8cm]{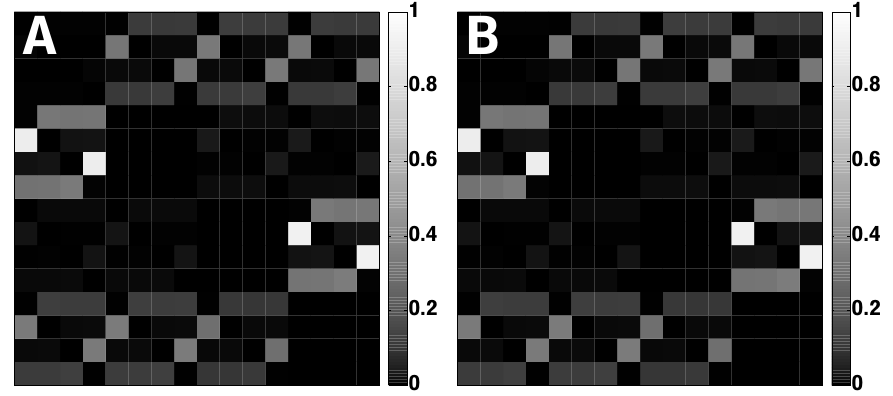}
\end{center}
\caption{\conf{1}{1} Policies.  A) Combined policy composed from two split
controllers (see text). B) Policy A after $10^6$ additional learning
iterations.}
\label{fig:combined from split}
\end{figure}

\begin{figure}[ht]
\begin{center}
  \includegraphics[width=16cm]{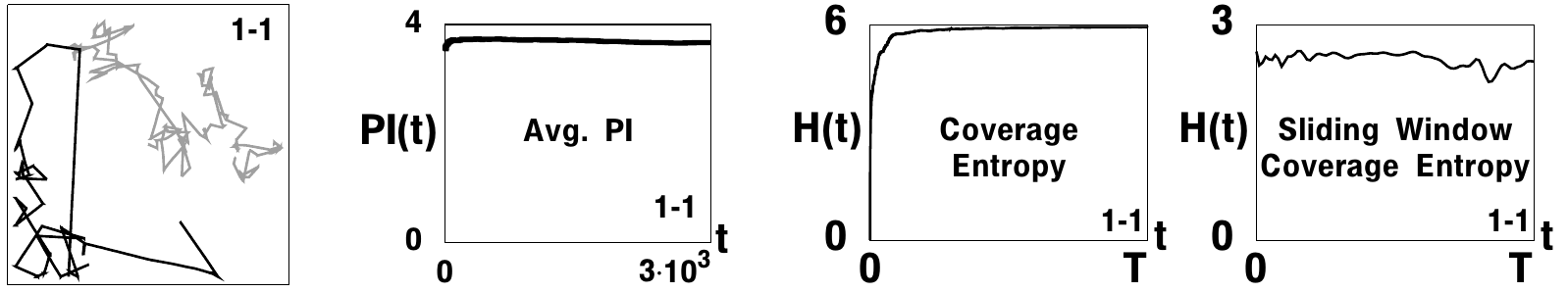}
\end{center}
\caption{From left to right: Trajectory, predictive information, coverage
entropy, sliding window coverage entropy. The plots show that the behavior is
unchanged, that the PI is maximal and doubled due to the doubling of the channel
capacity, and that the exploration behavior is equivalent to that of the split
controller configuration \conf{1}{2} (compare with Fig.~\ref{fig:mi plots} [1-2],
Fig.~\ref{fig:trajectory} [1-2], Fig.~\ref{fig:coverage entropy}).}
\label{fig:combined from split plots}
\end{figure}

Figure \ref{fig:combined from split}A shows the combined
controller, composed from the two optimal split controllers of the \conf{1}{2}
configuration (see Fig.~\ref{fig:1S}).
Using this matrix as an initialization for the learning process of the
\conf{1}{1} configuration leads to the policy shown in Figure \ref{fig:combined
from split}B. It must be noted that the learning process is now not anymore
restricted to the lower dimensional policy space of the split controllers, and
therefore, allows for further adjustments. However, it turns out that there is
no significant difference between the initial policy A and the converged policy
B with respect to the $L^2$-norm, which is 
$d = \sqrt{\sum_{i,j} (a_{ij} - b_{ij})^2} = 1.9 \cdot 10^{-6}$.
Consequently, the plots in Figure
\ref{fig:combined from split plots} show that the behavior has also not changed
significantly (compare with Fig.~\ref{fig:mi plots} [1-2],
Fig.~\ref{fig:trajectory} [1-2], Fig.~\ref{fig:coverage entropy})

This is an interesting result, as it shows that the optimal
solution in the sub-manifold of the split controllers is also an optimal
solution in the space of the combined controllers.
A geometric interpretation
of this result will follow in the discussion. This shows that a common
problem in learning agents, known as bootstrapping
\citep{Nolfi2000Evolutionary-Robotics}, which also occurs here in the case of the
\conf{1}{1} configuration, can be avoided using the same common strategy of
incremental learning for information maximization in the sensorimotor
loop.

To conclude this section, the derived learning rule is able to maximize the
predictive information for systems in the sensorimotor loop. Additionally, increases of
the PI relate to changes in the behavior and here to a higher coverage entropy,
an indirect measure for coordination among the coupled robots. 

\section{Discussion}
\label{sec:discussion}
This work presented a novel approach to self-organized learning in the
sensorimotor loop, which is free of assumptions on the world and restrictions
on the model. A learning algorithm was derived from the principle of
maximizing the (approximated) predictive information. Following the embodied
artificial intelligence approach, the learning rule was tested in experiments
with simulated robot chains. As desired, the average approximated predictive
information increased over time in each of the presented settings, which was the
primary goal of this work.

An important point here is that the increase of the average predictive
information alone does not allow conclusions to be drawn about 
specific changes of the robots behaviors. It is vital to relate the 
predictive information values to observations of the behaviors of the systems
while they interact in their environment, as it is the embodiment, which
determines the behavioral changes. This is an essential statement of
embodied artificial intelligence, and the reason why we chose robot experiments
to evaluate the learning rule.

The second result of this work is especially interesting because it is
counterintuitive.
The experiments show that there is a higher coverage entropy, a measure for
coordinated behavior in this setting, for chain configurations with more robots
and as well as such with split controllers. For three reasons this result is
counterintuitive;
1) each robot cannot measure the actions of the other robots in the chain directly,
but only through its wheel velocity sensor(s),
2) more robots mean that there is more disturbance in the motor-sensor coupling,
due to the higher physical interactions, which are a direct result of the
number of robots, and
3) the smaller controller setting can only read one sensor and has fewer
internal states, which makes it less capable of compensating for the
higher disturbances. 

We believe that the reason why less complex controllers coordinate better in
this setup can be well discussed in the context of morphological computation
\citep{Pfeifer2006How-the-Body}, but at this point we propose an
information-geometric approach. The set of split policies clearly forms a
low-dimensional subfamily of the family of all policies. Therefore, policies
with maximal predictive information should be reachable in that larger set. One
would even expect that the optimization in the set of split controllers is too
restrictive. However, the simulation results suggest that the split controllers
have a distinguished geometric property with regard to predictive information.
It seems that they constrain the optimization process in such a way that the
convergence towards a sub-optimal value of the predictive information becomes
unlikely. There is a huge number of local but not global maximizers with
moderate predictive information which are avoided through the constraints given
by split controllers. Furthermore, the policies that are reached by our learning
rule in most cases have a very high predictive information value. In summary,
bad policies are excluded and good policies are included through the particular
geometry of the family of split controllers. The geometric picture that we
sketched here, although not verified analytically, identifies selection criteria
for models (families of policies) in artificial learning systems based on the
maximization of objective functions such as predictive information. 

\appendix

\section*{Appendix}
\section{Derivation of the learning rule}
\label{sec:appendix derivation}
In order to derive a learning rule for the maximization of predictive
information we use the natural gradient method
\citep{Amari1998Natural-Gradient-Works},
which is based on the Fisher metric.
The application of this method to the policy context, within reinforcement
learning, has already been introduced by
\citep{Peters2005natural-actor-critic,Kakade2002A-natural-policy}.
In our case, where the
optimization domain, i.e.~the set of all policies, is given in terms of the
particular ``coordinate system,'' the gradient equations with respect to the
Fisher metric have the simple structure of replicator equations \cite{Hofbauer2003Evolutionary-game-dynamics}. Being
more precise, we denote by ${\mathcal P}({\mathcal{X}})$ the set of strictly
positive probability distributions on a non-empty finite set ${\mathcal{X}}$ and
consider a differentiable function $F: {\mathcal P}({\mathcal{X}}) \to \mathbb{R}$. With the gradient $\mathsf{grad}_p F$ of that function with respect to the
Fisher metric, which is also known as Shahshahani metric, the following
replicator equations are obtained (see also theorem 19.5.1 in
\citep{Hofbauer2003Evolutionary-game-dynamics}):
\[
\dot{p}(x) \; = \; 
\mathsf{grad}_p F (x) \; = \; 
p(x) \left( \partial_x F(p) - \sum_{x' \in {\mathcal{X}}} p(x') \, \partial_{x'} F(p) \right),          
\qquad  x \in {\mathcal{X}}. 
\]           
The right-hand side of the replicator equation is the gradient that we need for
the time discrete gradient ascent.
(Here we use $\partial_x F$ as the abbreviation for the partial derivative $\frac{\partial F}{\partial p(x)}$.)   
This gives us the update rule
\begin{equation} \label{iteration}
p^{(n+1)}(x) \; = \; p^{(n)}(x) + \frac{1}{n+1}  \,
                   p^{(n)}(x) \left( {\partial_x F}(p^{(n)}) - 
                   \sum_{x'} p^{(n)}(x') \, {\partial_{x'} F} (p^{(n)})
                   \right), \qquad x \in {\mathcal{X}}.
\end{equation}
We have chosen the rate $\frac{1}{n+1}$ in line with the general stochastic
approximation theory, where a typical assumption for the learn rates $a_n$ is
$\sum_{n=1}^\infty a_n = \infty$, $\sum_{n=1}^\infty a^2_n < \infty$
\citep{Benveniste1990Adaptive-algorithms-and}.
Now, after having outlined the general procedure, we come to the actual problem of maximizing predictive 
information. For each sensor value $s$ we consider as the optimization domain,
the space of policies $\alpha(a | s)$ which consists of probability
distributions on the set of actuator values. As derivative of the mutual
information $I(S';S)$ with respect to the policy $\alpha(a | s)$ we have:
\begin{align}
\frac{\partial I(S' ; S)}{\partial \alpha(a | s)} & =  p(s) \sum_{s'} \delta(s' | s,a) 
\ln \frac{\sum_a \alpha(a | s) \, \delta(s' | s,a)}{\sum_{s''}p(s'') \sum_a \alpha(a | s'') \, \delta(s' | s'',a)}
\label{eq:derivative}
\end{align}
Note that there is an implicit dependence of the stationary distribution $p(s)$
on the policy $\alpha(a|s)$ which complicates the derivative. This dependence is
not considered here, as it is subject of current research. 

Together with the general iteration rule (\ref{iteration}) the derivative
(\ref{eq:derivative}) results in a
corresponding iteration rule for the mutual information which almost coincides
with rule (\ref{eq:mil policy}). In order to obtain the
final step, note that the mutual information with respect to the kernel
$\delta(s' | s,a)$ is not necessarily consistent with the actual mutual
information generated through the mechanisms of the world. Therefore, we adjust
$\delta(s' | s,a)$ to the empirical data according to our rule (\ref{eq:mil world
model}). The
resulting sequence $\delta^{(n)}(s' | s,a)$ is then used for iteration
(\ref{iteration}).  

\subsubsection*{Acknowledgments}
We thank the anonymous referees for the constructive comments and 
Daniel Polani for the many fruitful discussions.


\newpage

\begin{thebibliography}{}

\bibitem[\protect\citeauthoryear{%
Amari%
}{%
Amari%
}{%
{\protect\APACyear{1998}}%
}]{%
Amari1998Natural-Gradient-Works}%
\APACinsertmetastar{%
Amari1998Natural-Gradient-Works}%
Amari, S.%
%
\newblock{}\BBOP{}1998\BBCP{}.
\newblock{}\BBOQ{}Natural gradient works efficiently in learning.\BBCQ{}
\newblock{}\Bem{Neural Computation}, \Bem{10}(2), 251-276.

\bibitem[\protect\citeauthoryear{%
Ay%
, Bertschinger%
, Der%
, G{\"u}ttler%
\BCBL{}\ \BBA{} Olbrich%
}{%
Ay%
\ \protect\BOthers{.}}{%
{\protect\APACyear{2008}}%
}]{%
Ay2008Predictive-information-and}%
\APACinsertmetastar{%
Ay2008Predictive-information-and}%
Ay, N.%
, Bertschinger, N.%
, Der, R.%
, G{\"u}ttler, F.%
\BCBL{}\ \BBA{} Olbrich, E.%
%
\newblock{}\BBOP{}2008\BBCP{}.
\newblock{}\BBOQ{}Predictive information and explorative behavior of autonomous
  robots.\BBCQ{}
\newblock{}\Bem{The European Physical Journal B - Condensed Matter and Complex
  Systems}, \Bem{63}(3), 329--339.

\bibitem[\protect\citeauthoryear{%
Barto%
}{%
Barto%
}{%
{\protect\APACyear{2004}}%
}]{%
Barto2004Intrinsically-motivated-learning}%
\APACinsertmetastar{%
Barto2004Intrinsically-motivated-learning}%
Barto, A.~G.%
%
\newblock{}\BBOP{}2004\BBCP{}.
\newblock{}\BBOQ{}Intrinsically motivated learning of hierarchical collections
  of skills.\BBCQ{}
\newblock{}\BIn{} \Bem{Proceedings of 3rd int. conference development learn.}\
  (\BPGS\ 112--119).
\newblock{}San Diego, CA, USA.

\bibitem[\protect\citeauthoryear{%
Benveniste%
, Priouret%
\BCBL{}\ \BBA{} M\'{e}tivier%
}{%
Benveniste%
\ \protect\BOthers{.}}{%
{\protect\APACyear{1990}}%
}]{%
Benveniste1990Adaptive-algorithms-and}%
\APACinsertmetastar{%
Benveniste1990Adaptive-algorithms-and}%
Benveniste, A.%
, Priouret, P.%
\BCBL{}\ \BBA{} M\'{e}tivier, M.%
%
\newblock{}\BBOP{}1990\BBCP{}.
\newblock{}\Bem{Adaptive algorithms and stochastic approximations}.
\newblock{}New York, NY, USA: Springer-Verlag New York, Inc.

\bibitem[\protect\citeauthoryear{%
Bertschinger%
}{%
Bertschinger%
}{%
{\protect\APACyear{2008}}%
}]{%
Bertschinger2008An-information-theoretic}%
\APACinsertmetastar{%
Bertschinger2008An-information-theoretic}%
Bertschinger, N.%
%
\newblock{}\BBOP{}2008\BBCP{}.
\newblock{}\Bem{An information theoretic perspective on cognitive systems:
  Memory and autonomy}.
\newblock{}\BPhD, Univeristy of Leipzig.

\bibitem[\protect\citeauthoryear{%
Bialek%
, Nemenman%
\BCBL{}\ \BBA{} Tishby%
}{%
Bialek%
\ \protect\BOthers{.}}{%
{\protect\APACyear{2001}}%
}]{%
Bialek2001Predictability-Complexity-and}%
\APACinsertmetastar{%
Bialek2001Predictability-Complexity-and}%
Bialek, W.%
, Nemenman, I.%
\BCBL{}\ \BBA{} Tishby, N.%
%
\newblock{}\BBOP{}2001\BBCP{}.
\newblock{}\BBOQ{}Predictability, complexity, and learning.\BBCQ{}
\newblock{}\Bem{Neural Computation}, \Bem{13}(11), 2409--2463.

\bibitem[\protect\citeauthoryear{%
Brooks%
}{%
Brooks%
}{%
{\protect\APACyear{1986}}%
}]{%
Brooks1986A-Robust-Layered}%
\APACinsertmetastar{%
Brooks1986A-Robust-Layered}%
Brooks, R.~A.%
%
\newblock{}\BBOP{}1986, March\BBCP{}.
\newblock{}\BBOQ{}A robust layered control system for a mobile robot.\BBCQ{}
\newblock{}\Bem{{IEEE} Journal of Robotics and Automation}, \Bem{2}(1), 14--23.

\bibitem[\protect\citeauthoryear{%
Brooks%
}{%
Brooks%
}{%
{\protect\APACyear{1991}}%
}]{%
Brooks1991Intelligence-Without-Reason}%
\APACinsertmetastar{%
Brooks1991Intelligence-Without-Reason}%
Brooks, R.~A.%
%
\newblock{}\BBOP{}1991\BBCP{}.
\newblock{}\BBOQ{}Intelligence without reason.\BBCQ{}
\newblock{}\BIn{} J.~Myopoulos\ \BBA{} R.~Reiter\ (\BEDS), \Bem{Proceedings of
  the 12th international joint conference on artificial intelligence
  ({IJCAI}-91)}\ (\BPGS\ 569--595).
\newblock{}Sydney, Australia: Morgan Kaufmann publishers Inc.: San Mateo, CA,
  USA.

\bibitem[\protect\citeauthoryear{%
Clark%
}{%
Clark%
}{%
{\protect\APACyear{1996}}%
}]{%
Clark1996Being-There:-Putting}%
\APACinsertmetastar{%
Clark1996Being-There:-Putting}%
Clark, A.%
%
\newblock{}\BBOP{}1996\BBCP{}.
\newblock{}\Bem{Being there: Putting brain, body, and world together again}.
\newblock{}Cambridge, MA, USA: MIT Press.

\bibitem[\protect\citeauthoryear{%
Cliff%
}{%
Cliff%
}{%
{\protect\APACyear{1990}}%
}]{%
Cliff1990Computational-neuroethology:-a}%
\APACinsertmetastar{%
Cliff1990Computational-neuroethology:-a}%
Cliff, D.%
%
\newblock{}\BBOP{}1990\BBCP{}.
\newblock{}\BBOQ{}Computational neuroethology: a provisional manifesto.\BBCQ{}
\newblock{}\BIn{} \Bem{Proceedings of the first international conference on
  simulation of adaptive behavior on from animals to animats}\ (\BPGS\ 29--39).
\newblock{}Cambridge, MA, USA: MIT Press.

\bibitem[\protect\citeauthoryear{%
Cover%
\ \BBA{} Thomas%
}{%
Cover%
\ \BBA{} Thomas%
}{%
{\protect\APACyear{2006}}%
}]{%
Cover2006Elements-of-Information}%
\APACinsertmetastar{%
Cover2006Elements-of-Information}%
Cover, T.~M.%
\BCBT{}\ \BBA{} Thomas, J.~A.%
%
\newblock{}\BBOP{}2006\BBCP{}.
\newblock{}\Bem{Elements of information theory}\ (2nd, \BED{}).
\newblock{}Wiley.

\bibitem[\protect\citeauthoryear{%
Crutchfield%
\ \BBA{} Young%
}{%
Crutchfield%
\ \BBA{} Young%
}{%
{\protect\APACyear{1989}}%
}]{%
Crutchfield1989Inferring-statistical-complexity}%
\APACinsertmetastar{%
Crutchfield1989Inferring-statistical-complexity}%
Crutchfield, J.~P.%
\BCBT{}\ \BBA{} Young, K.%
%
\newblock{}\BBOP{}1989, Jul\BBCP{}.
\newblock{}\BBOQ{}Inferring statistical complexity.\BBCQ{}
\newblock{}\Bem{Phys. Rev. Lett.}, \Bem{63}(2), 105--108.

\bibitem[\protect\citeauthoryear{%
Der%
}{%
Der%
}{%
{\protect\APACyear{2001}}%
}]{%
Der2001Self-Organized-Acquisition-of}%
\APACinsertmetastar{%
Der2001Self-Organized-Acquisition-of}%
Der, R.%
%
\newblock{}\BBOP{}2001\BBCP{}.
\newblock{}\BBOQ{}Self-organized acquisition of situated behavior.\BBCQ{}
\newblock{}\Bem{Theory in Biosciences}, \Bem{120}, 179-187.

\bibitem[\protect\citeauthoryear{%
Der%
, G{\"u}ttler%
\BCBL{}\ \BBA{} Ay%
}{%
Der%
\ \protect\BOthers{.}}{%
{\protect\APACyear{2008}}%
}]{%
Der2008Predictive-information-and}%
\APACinsertmetastar{%
Der2008Predictive-information-and}%
Der, R.%
, G{\"u}ttler, F.%
\BCBL{}\ \BBA{} Ay, N.%
%
\newblock{}\BBOP{}2008\BBCP{}.
\newblock{}\BBOQ{}Predictive information and emergent cooperativity in a chain
  of mobile robots.\BBCQ{}
\newblock{}\BIn{} \Bem{Alifexi proceedings.}
\newblock{}

\bibitem[\protect\citeauthoryear{%
Der%
\ \BBA{} Liebscher%
}{%
Der%
\ \BBA{} Liebscher%
}{%
{\protect\APACyear{2002}}%
}]{%
Der2002True-autonomy-from}%
\APACinsertmetastar{%
Der2002True-autonomy-from}%
Der, R.%
\BCBT{}\ \BBA{} Liebscher, R.%
%
\newblock{}\BBOP{}2002\BBCP{}.
\newblock{}\BBOQ{}True autonomy from self-organized adaptivity.\BBCQ{}
\newblock{}\BIn{} \Bem{Proc. workshop biologically inspired robotics. the
  legacy of grey walter 14-16 august 2002, bristol labs.}
\newblock{}Bristol.

\bibitem[\protect\citeauthoryear{%
Di~Paolo%
}{%
Di~Paolo%
}{%
{\protect\APACyear{2000}}%
}]{%
Di-Paolo2000Homeostatic-adaptation-to}%
\APACinsertmetastar{%
Di-Paolo2000Homeostatic-adaptation-to}%
Di~Paolo, E.~A.%
%
\newblock{}\BBOP{}2000\BBCP{}.
\newblock{}\BBOQ{}Homeostatic adaptation to inversion of the visual field and
  other sensorimotor disruptions.\BBCQ{}
\newblock{}\BIn{} J.-A. Meyer, A.~Berthoz, H.~Floreano D. and.~Roitblat\BCBL{}\
  \BBA{} S.~Wilson\ (\BEDS), \Bem{From animals to animats 6. proceedings of the
  {VI} international conference on simulation of adaptive behavior.}
\newblock{}Cambridge, MA: MIT Press.

\bibitem[\protect\citeauthoryear{%
F{\"o}rster%
}{%
F{\"o}rster%
}{%
{\protect\APACyear{1993}}%
}]{%
Forster1993Wissen-und-Gewissen}%
\APACinsertmetastar{%
Forster1993Wissen-und-Gewissen}%
F{\"o}rster, H. von.%
%
\newblock{}\BBOP{}1993\BBCP{}.
\newblock{}\Bem{{Wissen und Gewissen : Versuch einer Br{\"u}cke}}\ (1. Aufl.\
  \BEd; S.~J. Schmidt, \BED{}).
\newblock{}Frankfurt am Main, D: Suhrkamp.

\bibitem[\protect\citeauthoryear{%
Grassberger%
}{%
Grassberger%
}{%
{\protect\APACyear{1986}}%
}]{%
Grassberger1986Toward-a-quantitative}%
\APACinsertmetastar{%
Grassberger1986Toward-a-quantitative}%
Grassberger, P.%
%
\newblock{}\BBOP{}1986, 09\BBCP{}.
\newblock{}\BBOQ{}Toward a quantitative theory of self-generated
  complexity.\BBCQ{}
\newblock{}\Bem{International Journal of Theoretical Physics}, \Bem{25}(9),
  907--938.

\bibitem[\protect\citeauthoryear{%
Hofbauer%
\ \BBA{} Sigmund%
}{%
Hofbauer%
\ \BBA{} Sigmund%
}{%
{\protect\APACyear{2003}}%
}]{%
Hofbauer2003Evolutionary-game-dynamics}%
\APACinsertmetastar{%
Hofbauer2003Evolutionary-game-dynamics}%
Hofbauer, J.%
\BCBT{}\ \BBA{} Sigmund, K.%
%
\newblock{}\BBOP{}2003\BBCP{}.
\newblock{}\BBOQ{}Evolutionary game dynamics.\BBCQ{}
\newblock{}\Bem{Bull. Amer. Math. Soc.}, \Bem{40}, 479--519.

\bibitem[\protect\citeauthoryear{%
Kakade%
}{%
Kakade%
}{%
{\protect\APACyear{2002}}%
}]{%
Kakade2002A-natural-policy}%
\APACinsertmetastar{%
Kakade2002A-natural-policy}%
Kakade, S.%
%
\newblock{}\BBOP{}2002\BBCP{}.
\newblock{}\BBOQ{}A natural policy gradient.\BBCQ{}
\newblock{}\Bem{Advances in neural information processing systems}, \Bem{2},
  1531--1538.

\bibitem[\protect\citeauthoryear{%
Kaplan%
\ \BBA{} Oudeyer%
}{%
Kaplan%
\ \BBA{} Oudeyer%
}{%
{\protect\APACyear{2004}}%
}]{%
Kaplan2004Maximizing-Learning-Progress:}%
\APACinsertmetastar{%
Kaplan2004Maximizing-Learning-Progress:}%
Kaplan, F.%
\BCBT{}\ \BBA{} Oudeyer, P.-Y.%
%
\newblock{}\BBOP{}2004\BBCP{}.
\newblock{}\BBOQ{}Maximizing learning progress: An internal reward system for
  development.\BBCQ{}
\newblock{}\Bem{Embodied Artificial Intelligence}, 629--629.

\bibitem[\protect\citeauthoryear{%
Laughlin%
}{%
Laughlin%
}{%
{\protect\APACyear{1981}}%
}]{%
Laughlin1981A-simple-coding}%
\APACinsertmetastar{%
Laughlin1981A-simple-coding}%
Laughlin, S.%
%
\newblock{}\BBOP{}1981\BBCP{}.
\newblock{}\BBOQ{}A simple coding procedure enhances a neuron's information
  capacity.\BBCQ{}
\newblock{}\Bem{Z Naturforsch C}, \Bem{36}(9-10), 910-2.

\bibitem[\protect\citeauthoryear{%
Linsker%
}{%
Linsker%
}{%
{\protect\APACyear{1988}}%
}]{%
Linsker1988Self-organization-in-a}%
\APACinsertmetastar{%
Linsker1988Self-organization-in-a}%
Linsker, R.%
%
\newblock{}\BBOP{}1988\BBCP{}.
\newblock{}\BBOQ{}Self-organization in a perceptual network.\BBCQ{}
\newblock{}\Bem{IEEE Computers}, \Bem{88}, 105--117.

\bibitem[\protect\citeauthoryear{%
Lungarella%
\ \BBA{} Sporns%
}{%
Lungarella%
\ \BBA{} Sporns%
}{%
{\protect\APACyear{2005}}%
}]{%
Lungarella2005Information-Self-Structuring:-Key}%
\APACinsertmetastar{%
Lungarella2005Information-Self-Structuring:-Key}%
Lungarella, M.%
\BCBT{}\ \BBA{} Sporns, O.%
%
\newblock{}\BBOP{}2005\BBCP{}.
\newblock{}\BBOQ{}Information self-structuring: Key principle for learning and
  development.\BBCQ{}
\newblock{}\Bem{Development and Learning, 2005. Proceedings. The 4th
  International Conference on}, 25--30.

\bibitem[\protect\citeauthoryear{%
Meltzoff%
\ \BBA{} Moore%
}{%
Meltzoff%
\ \BBA{} Moore%
}{%
{\protect\APACyear{1997}}%
}]{%
Meltzoff1997Explaining-facial-imitation:}%
\APACinsertmetastar{%
Meltzoff1997Explaining-facial-imitation:}%
Meltzoff, A.%
\BCBT{}\ \BBA{} Moore, M.~K.%
%
\newblock{}\BBOP{}1997\BBCP{}.
\newblock{}\BBOQ{}Explaining facial imitation: A theoretical model.\BBCQ{}
\newblock{}\Bem{Early Development and Parenting}, 179--192.

\bibitem[\protect\citeauthoryear{%
Mondada%
, Franzi%
\BCBL{}\ \BBA{} Ienne%
}{%
Mondada%
\ \protect\BOthers{.}}{%
{\protect\APACyear{1993}}%
}]{%
Mondada1993Mobile-Robot-Miniaturization:}%
\APACinsertmetastar{%
Mondada1993Mobile-Robot-Miniaturization:}%
Mondada, F.%
, Franzi, E.%
\BCBL{}\ \BBA{} Ienne, P.%
%
\newblock{}\BBOP{}1993\BBCP{}.
\newblock{}\BBOQ{}Mobile robot miniaturization: A tool for investigation in
  control algorithms.\BBCQ{}
\newblock{}\BIn{} \Bem{Proceedings of the third international symposium on
  experimental robotics}\ (\BPGS\ 501--513).
\newblock{}Berlin: Springer Verlag.

\bibitem[\protect\citeauthoryear{%
Nolfi%
\ \BBA{} Floreano%
}{%
Nolfi%
\ \BBA{} Floreano%
}{%
{\protect\APACyear{2000}}%
}]{%
Nolfi2000Evolutionary-Robotics}%
\APACinsertmetastar{%
Nolfi2000Evolutionary-Robotics}%
Nolfi, S.%
\BCBT{}\ \BBA{} Floreano, D.%
%
\newblock{}\BBOP{}2000\BBCP{}.
\newblock{}\Bem{Evolutionary robotics}.
\newblock{}MIT Press.

\bibitem[\protect\citeauthoryear{%
Oudeyer%
, Kaplan%
\BCBL{}\ \BBA{} Hafner%
}{%
Oudeyer%
\ \protect\BOthers{.}}{%
{\protect\APACyear{2007}}%
}]{%
Oudeyer2007Intrinsic-Motivation-Systems}%
\APACinsertmetastar{%
Oudeyer2007Intrinsic-Motivation-Systems}%
Oudeyer, P.-Y.%
, Kaplan, F.%
\BCBL{}\ \BBA{} Hafner, V.%
%
\newblock{}\BBOP{}2007, April\BBCP{}.
\newblock{}\BBOQ{}Intrinsic motivation systems for autonomous mental
  development.\BBCQ{}
\newblock{}\Bem{Evolutionary Computation, IEEE Transactions on}, \Bem{11}(2),
  265-286.

\bibitem[\protect\citeauthoryear{%
Pasemann%
, Zahedi%
\BCBL{}\ \BBA{} Rohde%
}{%
Pasemann%
\ \protect\BOthers{.}}{%
{\protect\APACyear{2004}}%
}]{%
Pasemann2004Adaptive-Behaviour-Control}%
\APACinsertmetastar{%
Pasemann2004Adaptive-Behaviour-Control}%
Pasemann, F.%
, Zahedi, K.%
\BCBL{}\ \BBA{} Rohde, M.%
%
\newblock{}\BBOP{}2004\BBCP{}.
\newblock{}\Bem{Adaptive behaviour control by self-regulating neurons}\
  (Preprint\ \BNUM~55).
\newblock{}MPI MiS.

\bibitem[\protect\citeauthoryear{%
Peters%
, Vijayakumar%
\BCBL{}\ \BBA{} Schaal%
}{%
Peters%
\ \protect\BOthers{.}}{%
{\protect\APACyear{2005}}%
}]{%
Peters2005natural-actor-critic}%
\APACinsertmetastar{%
Peters2005natural-actor-critic}%
Peters, J.%
, Vijayakumar, S.%
\BCBL{}\ \BBA{} Schaal, S.%
%
\newblock{}\BBOP{}2005\BBCP{}.
\newblock{}\BBOQ{}Natural actor-critic.\BBCQ{}
\newblock{}\BIn{} \Bem{proceedings of the 16th european conference on machine
  learning (ecml 2005)}\ (\BPG\ 280-291).
\newblock{}springer.

\bibitem[\protect\citeauthoryear{%
Pfeifer%
\ \BBA{} Bongard%
}{%
Pfeifer%
\ \BBA{} Bongard%
}{%
{\protect\APACyear{2006}}%
}]{%
Pfeifer2006How-the-Body}%
\APACinsertmetastar{%
Pfeifer2006How-the-Body}%
Pfeifer, R.%
\BCBT{}\ \BBA{} Bongard, J.~C.%
%
\newblock{}\BBOP{}2006\BBCP{}.
\newblock{}\Bem{How the body shapes the way we think: A new view of
  intelligence}.
\newblock{}The MIT Press (Bradford Books).

\bibitem[\protect\citeauthoryear{%
Pfeifer%
, Lungarella%
\BCBL{}\ \BBA{} Iida%
}{%
Pfeifer%
\ \protect\BOthers{.}}{%
{\protect\APACyear{2007}}%
}]{%
Pfeifer2007Self-Organization-Embodiment-and}%
\APACinsertmetastar{%
Pfeifer2007Self-Organization-Embodiment-and}%
Pfeifer, R.%
, Lungarella, M.%
\BCBL{}\ \BBA{} Iida, F.%
%
\newblock{}\BBOP{}2007\BBCP{}.
\newblock{}\BBOQ{}Self-organization, embodiment, and biologically inspired
  robotics.\BBCQ{}
\newblock{}\Bem{Science}, \Bem{318}(5853), 1088--1093.

\bibitem[\protect\citeauthoryear{%
Polani%
, Nehaniv%
, Martinetz%
\BCBL{}\ \BBA{} Kim%
}{%
Polani%
\ \protect\BOthers{.}}{%
{\protect\APACyear{2006}}%
}]{%
Polani2006Relevant-Information-in}%
\APACinsertmetastar{%
Polani2006Relevant-Information-in}%
Polani, D.%
, Nehaniv, C.%
, Martinetz, T.%
\BCBL{}\ \BBA{} Kim, J.~T.%
%
\newblock{}\BBOP{}2006\BBCP{}.
\newblock{}\BBOQ{}Relevant {I}nformation in {O}ptimized {P}ersistence vs.
  {P}rogeny {S}trategies.\BBCQ{}
\newblock{}\BIn{} L.~M.Rocha, M.~Bedau, D.~Floreano, R.~Goldstone,
  A.~Vespignani\BCBL{}\ \BBA{} L.~Yaeger\ (\BEDS), \Bem{Proc. artificial life
  x.}
\newblock{}

\bibitem[\protect\citeauthoryear{%
Porr%
}{%
Porr%
}{%
{\protect\APACyear{2003}}%
}]{%
Porr2003Sequence-Learning-in-a}%
\APACinsertmetastar{%
Porr2003Sequence-Learning-in-a}%
Porr, B.%
%
\newblock{}\BBOP{}2003\BBCP{}.
\newblock{}\Bem{Sequence-learning in a self-referential closed-loop behavioural
  system}.
\newblock{}\BPhD, Faculty of Human Sciences, Department of Psychology.

\bibitem[\protect\citeauthoryear{%
Schmidhuber%
}{%
Schmidhuber%
}{%
{\protect\APACyear{1990}}%
}]{%
Schmidhuber1990A-possibility-for}%
\APACinsertmetastar{%
Schmidhuber1990A-possibility-for}%
Schmidhuber, J.%
%
\newblock{}\BBOP{}1990\BBCP{}.
\newblock{}\BBOQ{}A possibility for implementing curiosity and boredom in
  model-building neural controllers.\BBCQ{}
\newblock{}\BIn{} \Bem{Proceedings of the first international conference on
  simulation of adaptive behavior on from animals to animats}\ (\BPGS\
  222--227).
\newblock{}Cambridge, MA, USA: MIT Press.

\bibitem[\protect\citeauthoryear{%
Schmidhuber%
}{%
Schmidhuber%
}{%
{\protect\APACyear{2009}}%
}]{%
Schmidhuber2009Driven-by-Compression}%
\APACinsertmetastar{%
Schmidhuber2009Driven-by-Compression}%
Schmidhuber, J.%
%
\newblock{}\BBOP{}2009\BBCP{}.
\newblock{}\BBOQ{}Driven by compression progress: A simple principle explains
  essential aspects of subjective beauty, novelty, surprise, interestingness,
  attention, curiosity, creativity, art, science, music, jokes.\BBCQ{}
\newblock{}\Bem{Anticipatory Behavior in Adaptive Learning Systems}, 48--76.

\bibitem[\protect\citeauthoryear{%
Shannon%
}{%
Shannon%
}{%
{\protect\APACyear{1948}}%
}]{%
Shannon1948A-mathematical-theory}%
\APACinsertmetastar{%
Shannon1948A-mathematical-theory}%
Shannon, C.~E.%
%
\newblock{}\BBOP{}1948\BBCP{}.
\newblock{}\BBOQ{}A mathematical theory of communication.\BBCQ{}
\newblock{}\Bem{Bell System Technical Journal}, \Bem{27}, 379---423.

\bibitem[\protect\citeauthoryear{%
Steels%
}{%
Steels%
}{%
{\protect\APACyear{2004}}%
}]{%
Steels2004The-Autotelic-Principle}%
\APACinsertmetastar{%
Steels2004The-Autotelic-Principle}%
Steels, L.%
%
\newblock{}\BBOP{}2004\BBCP{}.
\newblock{}\BBOQ{}The autotelic principle.\BBCQ{}
\newblock{}\Bem{Embodied Artificial Intelligence}, 629--629.

\bibitem[\protect\citeauthoryear{%
Steels%
\ \BBA{} Wellens%
}{%
Steels%
\ \BBA{} Wellens%
}{%
{\protect\APACyear{2007}}%
}]{%
Steels2007Scaffolding-Language-Emergence}%
\APACinsertmetastar{%
Steels2007Scaffolding-Language-Emergence}%
Steels, L.%
\BCBT{}\ \BBA{} Wellens, P.%
%
\newblock{}\BBOP{}2007\BBCP{}.
\newblock{}\BBOQ{}Scaffolding language emergence using the autotelic
  principle.\BBCQ{}
\newblock{}\BIn{} \Bem{Ieee symposium on artificial life}\ (\BPG\ 325-332).
\newblock{}

\bibitem[\protect\citeauthoryear{%
Still%
}{%
Still%
}{%
{\protect\APACyear{2009}}%
}]{%
Still2009Information-theoretic-approach-to}%
\APACinsertmetastar{%
Still2009Information-theoretic-approach-to}%
Still, S.%
%
\newblock{}\BBOP{}2009, jan\BBCP{}.
\newblock{}\BBOQ{}Information-theoretic approach to interactive
  learning.\BBCQ{}
\newblock{}\Bem{EPL}, \Bem{85}(2), 28005.

\bibitem[\protect\citeauthoryear{%
Storck%
, Hochreiter%
\BCBL{}\ \BBA{} Schmidhuber%
}{%
Storck%
\ \protect\BOthers{.}}{%
{\protect\APACyear{1995}}%
}]{%
Storck1995Reinforcement-Driven-Information}%
\APACinsertmetastar{%
Storck1995Reinforcement-Driven-Information}%
Storck, J.%
, Hochreiter, S.%
\BCBL{}\ \BBA{} Schmidhuber, J.%
%
\newblock{}\BBOP{}1995\BBCP{}.
\newblock{}\BBOQ{}Reinforcement driven information acquisition in
  non-deterministic environments.\BBCQ{}
\newblock{}\BIn{} \Bem{Proceedings of the international conference on
  artificial neural networks}\ (\BPGS\ 159--164).
\newblock{}

\bibitem[\protect\citeauthoryear{%
Thelen%
\ \BBA{} Smith%
}{%
Thelen%
\ \BBA{} Smith%
}{%
{\protect\APACyear{1996}}%
}]{%
Thelen1996A-dynamic-systems}%
\APACinsertmetastar{%
Thelen1996A-dynamic-systems}%
Thelen, E.%
\BCBT{}\ \BBA{} Smith, L.~B.%
%
\newblock{}\BBOP{}1996\BBCP{}.
\newblock{}\Bem{A dynamic systems approach to the development of cognition and
  action}.
\newblock{}Cambridge, Mass. {$[$}u.a.{$]$}: MIT Press.

\bibitem[\protect\citeauthoryear{%
Uexkuell%
}{%
Uexkuell%
}{%
{\protect\APACyear{1957 [1934]}}%
}]{%
Uexkuell1957A-Stroll-Through}%
\APACinsertmetastar{%
Uexkuell1957A-Stroll-Through}%
Uexkuell, J. von.%
%
\newblock{}\BBOP{}1957 [1934]\BBCP{}.
\newblock{}\BBOQ{}A stroll through the worlds of animals and men.\BBCQ{}
\newblock{}\BIn{} C.~H. Schiller\ (\BED), \Bem{Instinctive behavior}\ (\BPG\
  5-80).
\newblock{}New York: International Universities Press.

\bibitem[\protect\citeauthoryear{%
Williams%
\ \BBA{} Kelley%
}{%
Williams%
\ \BBA{} Kelley%
}{%
{\protect\APACyear{2009}}%
}]{%
Williams2009gnuplot-4.2.6}%
\APACinsertmetastar{%
Williams2009gnuplot-4.2.6}%
Williams, T.%
\BCBT{}\ \BBA{} Kelley, C.%
%
\newblock{}\BBOP{}2009, September\BBCP{}.
\newblock{}\Bem{gnuplot 4.2.6.}
\newblock{}\url{http://www.gnuplot.info}.
\newblock{}

\bibitem[\protect\citeauthoryear{%
Zahedi%
, Twickel%
\BCBL{}\ \BBA{} Pasemann%
}{%
Zahedi%
\ \protect\BOthers{.}}{%
{\protect\APACyear{2008}}%
}]{%
Zahedi2008YARS:-A-Physical}%
\APACinsertmetastar{%
Zahedi2008YARS:-A-Physical}%
Zahedi, K.%
, Twickel, A. von%
\BCBL{}\ \BBA{} Pasemann, F.%
%
\newblock{}\BBOP{}2008\BBCP{}.
\newblock{}\BBOQ{}Yars: A physical 3d simulator for evolving controllers for
  real robots.\BBCQ{}
\newblock{}\BIn{} S.~Carpin\ \BBA{} et al.\ (\BEDS), \Bem{Simpar 2008}\ (\BPGS\
  71---82).
\newblock{}Springer.

\end{thebibliography}
\end{document}